\documentclass{elsarticle}\usepackage[]{graphicx}\usepackage[]{color}
\makeatletter
\def\maxwidth{ %
  \ifdim\Gin@nat@width>\linewidth
    \linewidth
  \else
    \Gin@nat@width
  \fi
}
\makeatother

\definecolor{fgcolor}{rgb}{0.345, 0.345, 0.345}
\newcommand{\hlnum}[1]{\textcolor[rgb]{0.686,0.059,0.569}{#1}}%
\newcommand{\hlstr}[1]{\textcolor[rgb]{0.192,0.494,0.8}{#1}}%
\newcommand{\hlcom}[1]{\textcolor[rgb]{0.678,0.584,0.686}{\textit{#1}}}%
\newcommand{\hlopt}[1]{\textcolor[rgb]{0,0,0}{#1}}%
\newcommand{\hlstd}[1]{\textcolor[rgb]{0.345,0.345,0.345}{#1}}%
\newcommand{\hlkwa}[1]{\textcolor[rgb]{0.161,0.373,0.58}{\textbf{#1}}}%
\newcommand{\hlkwb}[1]{\textcolor[rgb]{0.69,0.353,0.396}{#1}}%
\newcommand{\hlkwc}[1]{\textcolor[rgb]{0.333,0.667,0.333}{#1}}%
\newcommand{\hlkwd}[1]{\textcolor[rgb]{0.737,0.353,0.396}{\textbf{#1}}}%

\usepackage{framed}
\makeatletter
\newenvironment{kframe}{%
 \def\at@end@of@kframe{}%
 \ifinner\ifhmode%
  \def\at@end@of@kframe{\end{minipage}}%
  \begin{minipage}{\columnwidth}%
 \fi\fi%
 \def\FrameCommand##1{\hskip\@totalleftmargin \hskip-\fboxsep
 \colorbox{shadecolor}{##1}\hskip-\fboxsep
     \hskip-\linewidth \hskip-\@totalleftmargin \hskip\columnwidth}%
 \MakeFramed {\advance\hsize-\width
   \@totalleftmargin\z@ \linewidth\hsize
   \@setminipage}}%
 {\par\unskip\endMakeFramed%
 \at@end@of@kframe}
\makeatother

\definecolor{shadecolor}{rgb}{.97, .97, .97}
\definecolor{messagecolor}{rgb}{0, 0, 0}
\definecolor{warningcolor}{rgb}{1, 0, 1}
\definecolor{errorcolor}{rgb}{1, 0, 0}
\newenvironment{knitrout}{}{} 

\usepackage{alltt}
\usepackage[utf8]{inputenc}
\usepackage[T1]{fontenc}
\usepackage[english]{babel}
\usepackage{lineno, hyperref}
\usepackage{amsmath, amsfonts,amssymb}
\usepackage{mathtools}
\usepackage{tabularx}
\usepackage{algorithm}
\usepackage{algpseudocode}

\usepackage{booktabs}
\usepackage{csquotes}

\usepackage{nicefrac}
\usepackage{siunitx}
\usepackage[inline,shortlabels]{enumitem}

\usepackage{tikz, xcolor}
\usetikzlibrary{shapes,arrows,positioning}

\usepackage{amsmath, amsfonts}
\usepackage{mathtools}
\usepackage{xspace}

\newcommand{\R}{\ensuremath{\mathbb{R}}}                   
\DeclareMathOperator*{\argmin}{arg\,min}                   
\DeclareMathOperator*{\Inf}{Inf}                           
\newcommand{\EI}{\ensuremath{\operatorname{EI}}\xspace}    

\newcommand{\X}{\ensuremath{\mathcal{X}}\xspace}           
\newcommand{\xv}{\ensuremath{\boldsymbol{x}}\xspace}       
\newcommand{\yv}{\ensuremath{\boldsymbol{y}}\xspace}       
\newcommand{\fh}{\ensuremath{\hat{f}}\xspace}              
\newcommand{\fx}{\ensuremath{f(\xv)}\xspace}               
\newcommand{\sh}{\ensuremath{\hat{s}}\xspace}              
\newcommand{\shx}{\ensuremath{\sh(\xv)}\xspace}            
\newcommand{\vhx}{\ensuremath{\sh^2(\xv)}\xspace}          
\newcommand{\mh}{\ensuremath{\hat{\mu}}\xspace}            
\newcommand{\mhx}{\ensuremath{\mh(\xv)}\xspace}            

\newcommand{\Rlang}{\texttt{R}\xspace}                     
\newcommand{\mlrMBO}{\textbf{\texttt{mlrMBO}}\xspace}      
\newcommand{\mlr}{\textbf{\texttt{mlr}}\xspace}            
\newcommand{\pkg}[1]{\texttt{#1}\xspace}                   
\newcommand{\MBO}{model-based optimization\xspace}         
\newcommand{\ie}{i.e.\xspace}                             
\newcommand{\eg}{e.g.\xspace}                             
\newcommand{\code}[1]{\texttt{#1}}

\usepackage{pifont}

\newcommand{\yh}{\ensuremath{\boldsymbol{\hat{y}}}\xspace} 

\journal{Computational Statistics \& Data Analysis}

\IfFileExists{upquote.sty}{\usepackage{upquote}}{}
\begin{document}

  \begin{frontmatter}

  \title{mlrMBO: A Modular Framework for Model-Based Optimization of Expensive Black-Box Functions}

  \address[lmu]{Ludwig-Maximilians-Universit\"at M\"unchen, Germany}
  \address[tudo]{TU Dortmund University, Germany}
  \address[wwu]{Westf\"alische-Wilhelms Universit\"at M\"unster, Germany}

  \author[lmu]{Bernd Bischl}
  \ead{bernd\_bischl@gmx.net}

  \author[tudo]{Jakob Richter}
  \ead{jakob.richter@tu-dortmund.de}

  \author[wwu]{Jakob Bossek}
  \ead{bossek@wi.uni-muenster.de}

  \author[tudo]{Daniel Horn}
  \ead{daniel.horn@tu-dortmund.de}

  \author[lmu]{Janek Thomas}
  \ead{janek.thomas@stat.uni-muenchen.de}

  \author[tudo]{Michel Lang}
  \ead{lang@statistik.tu-dortmund.de}

\begin{abstract}
   We present \mlrMBO, a flexible and comprehensive \Rlang toolbox for \MBO~(MBO), also known as Bayesian optimization, which addresses the problem of expensive black-box optimization by approximating the given objective function through a surrogate regression model.
   It is designed for both single- and multi-objective optimization with mixed continuous, categorical and conditional parameters.
   Additional features include multi-point batch proposal, parallelization, visualization, logging and error-handling.
   \mlrMBO is implemented in a modular fashion, such that single components can be easily replaced or adapted by the user for specific use cases, e.g., any regression learner from the \texttt{mlr} toolbox for machine learning can be used, and infill criteria and infill optimizers are easily exchangeable.
   We empirically demonstrate that \mlrMBO provides state-of-the-art performance by comparing it on different benchmark scenarios against a wide range of other optimizers, including DiceOptim, rBayesianOptimization, SPOT, SMAC, Spearmint, and Hyperopt.
\end{abstract}


  \end{frontmatter}


\section{Introduction}

\label{sec:introduction}
Black-box functions are systems that require a number of input parameters to produce one or multiple (numeric) outputs.
In most cases these are
\begin{enumerate*}[(a)]
  \item expensive to evaluate in terms of time and/or monetary cost, and
  \item knowledge of their internal working is not available, which often manifests through the absence of derivatives.
\end{enumerate*}
Such problems occur in production engineering~\cite[\eg][]{Sieben2010}, where the inputs are possible settings of industrial machines or used materials and the output is one or multiple measurements regarding the quality of fabricated parts.
Since this makes a single evaluation expensive, one tries to find the optimal settings of production steps in a minimal number of tries. 
Design of Computer Experiments (DACE) \cite{SWMW89} is a discipline focused on solving such problems and sequential model-based optimization (SMBO)~\cite{jones_1998} has become the state-of-the-art optimization strategy in recent years.

The generic SMBO procedure starts with an initial design of evaluation points, and then iterates the following steps:
\begin{enumerate}
  \item Fit a regression model to the outcomes and design points obtained so far,
  \item query the model to propose a new, promising point, often by optimizing a so-called infill criterion or acquisition function,
  \item evaluate the new point with the black-box function and add it to the design.
\end{enumerate}
Several adaptations and extensions, \eg, multi-objective optimization~\cite{HWB15}, multi-point proposal~\cite{ginsbourger2010kriging,bischl_2014}, more flexible regression models~\cite{hutter_sequential_2011} or alternative ways to calculate the infill criterion~\cite{bergstra2011algorithms} have been investigated recently.

A different field of application for SMBO is the hyperparameter optimization for machine learning methods~\cite[\eg][]{thornton_2013,lang_2015,HB2016}.
Here, the black-box is a machine learning method and the objective(s) is one or multiple performance measure(s), validated via resampling on a data set of interest.
The black-box function can be more complex, for example a machine learning pipeline which includes preprocessing, feature selection and model selection.

\medskip
After a brief comparison with related software in Subsection~\ref{ssec:related} and clarification of our main contributions in Subsection~\ref{ssec:contributions}, we introduce the general SMBO procedure in more detail in Section~\ref{sec:methodology}.
Section~\ref{sec:software} highlights the capabilities of our software \mlrMBO, showcased by some code examples.
In Section~\ref{sec:benchmarks} we empirically demonstrate that \mlrMBO achieves state-of-the-art performance on a wide range of synthetic and real-world single- and multi-objective scenarios.
Section~\ref{sec:conclusion} gives an outlook on future work.

\subsection{Related Software}
\label{ssec:related}

We will briefly present an overview of available software for \MBO, starting with implementations based on the Efficient Global Optimization algorithm (EGO), \ie, the SMBO algorithm proposed by \citet{jones_1998} using Gaussian processes (GPs), and continue with extensions and alternative approaches.

Both \pkg{DiceOptim}~\cite{RGD12} and \pkg{rBayesianOptimization}~\cite{R:rBayesianOptimization} are \Rlang packages that offer EGO implementations.
A sophisticated EGO implementation can be found in the Python package \pkg{Spearmint}~\cite{snoek_practical_2012}.
It focuses on hyperparameter optimization of machine learning algorithms with enhancements regarding variable costs of experiments and parallelization.
All three packages offer different GP kernels and infill criteria, but only support numerical (non-conditional) parameters and, except for Spearmint, no multi-criteria optimization or parallelization is available. A multi-criteria version of Spearmint is introduced in \cite{hernandez2016predictive}.

The C++ library \pkg{BayesOpt}~\cite{JMLR:v15:martinezcantin14a} contains an extended version of EGO, including Student-t processes, support of mixed and conditional parameters as well as meta-criteria algorithms to automatically find reasonable infill criteria during optimization.
It offers interfaces for Python, Matlab and Octave.

\pkg{SMAC}~\cite{hutter_sequential_2011} is one of the most established frameworks and allows to optimize mixed parameter spaces as it uses a random forest instead of a GP for regression.
Besides general black-box optimization, it is focused on algorithm configuration.
However, \pkg{SMAC} is limited to single-criteria optimization and parallelization is not supported.

\pkg{Hyperopt}~\cite{bergstra2011algorithms} is an optimization package in Python that supports numerical, categorical and conditional parameters.
Instead of a regression it uses a tree of Parzen estimators (TPE) to compute point suggestions.
It supports distributed parallel and asynchronous execution.
\pkg{Hyperopt} can be used for general black-box optimization, but is mainly focused on machine learning tasks.

Another \Rlang implementation for sequential black-box optimization is \pkg{SPOT}~\cite{beielstein_spot_2012}.
It is a toolbox with different modeling techniques and offers a wide variety of statistical methods.
\pkg{SPOT} contains sophisticated algorithms to handle functions with noisy evaluations, is able to handle constrains in functions and supports multi-objective optimization.


\subsection{Main Contributions and Prior Applications}
\label{ssec:contributions}

The main contribution of this paper is the presentation of the \Rlang package \mlrMBO, which implements a generic SMBO framework and provides a large variety of different SMBO methods due to its modular structure.
\mlrMBO is even more flexible than SPOT in its choice of surrogate models as it is connected to the \Rlang package \mlr~\cite{mlr_2016} which interfaces more than $60$ machine learning regression algorithms.
Besides the default SMBO procedure, \mlrMBO focuses on three domains: Mixed parameter space optimization, multi-point proposals and multi-objective optimization.
Even combinations of the three domains are possible, which to our knowledge no other software is currently capable of.
\mlrMBO is easy to use as many default implementations for the individual steps of the SMBO procedure are directly supported in a plug-and-play style.
Simple interfaces are available to extend the package with user specific variants.

Benchmarks show that \mlrMBO achieves state-of-the-art performance in each domain.
Additionally, \mlrMBO has been successfully applied in some practical settings.
In \cite{koch2012,bischl2014} it was used to optimize the hyperparameters of machine learning pipelines (joint pre-processing and model hyperparameters) for support vector machines and general machine learning models, respectively, in a single objective setting.
\citet{hess2013} proposed an \mlrMBO ensemble-based approach to identify the best surrogate model during optimization through reinforcement learning.
\citet{HDBGW2016} considered a multi-objective benchmark and optimized the runtime-accuracy trade-offs of several approximate support vector machine solvers.
\citet{HB2016} introduced the general capability of \mlrMBO to solve multi-objective machine learning tasks.
\citet{Steponavice2016} investigated the impact of different initial design sampling techniques on the performance of multi-objective model-based optimization methods by using \mlrMBO.

\section{Sequential Model-Based Optimization}
\label{sec:methodology}
This section describes the general SMBO setup and presents the individual building blocks in Subsection~\ref{ssec:smbo}.
While SMBO is modular and can thus be customized for a variety of
different tasks, we highlight the most prominent combinations of components
described in the literature like EGO~\cite{jones_1998}
(Subsection~\ref{ssec:ego}) or SMAC-like~\cite{hutter_sequential_2011} optimizers (Subsection~\ref{ssec:mso}).
Subsections~\ref{ssec:multipoint} and~\ref{ssec:multiobjective} introduce parallelization through multi-point proposal, and multi-objective optimization.

\subsection{Sequential model-based optimization}
\label{ssec:smbo}
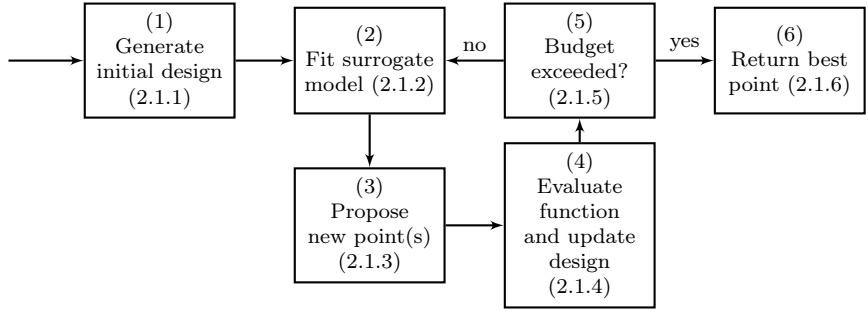
\begin{figure}[htb]
  \begin{center}

\tikzstyle{b} = [rectangle, draw, node distance=2em and 2.2em, text width=5em, text centered, minimum height=4em, thick, font=\footnotesize]
\tikzstyle{l} = [draw, -latex',thick]

\begin{tikzpicture}[auto]
  \node [b] (design) {(1)\\ Generate initial design (\ref{sssec:initial})};
    \node [rectangle, left=of design] (off) {};
    \node [b, right=of design] (fitting) {(2)\\ Fit surrogate model (\ref{sssec:surrogate})};
    \node [b, right=of fitting] (budget) {(5)\\ Budget exceeded?  (\ref{sssec:termination})};
    \node [b, right=of budget] (return) {(6)\\ Return best point (\ref{sssec:final_point})};
    \node [b, below=of fitting] (infill) {(3)\\ Propose new point(s) (\ref{sssec:infill})};
    \node [b, right=of infill] (update) {(4)\\ Evaluate function and update design (\ref{sssec:infill_opt})};

    \path [l] (off) -- (design);
    \path [l] (design) -- (fitting);
    \path [l] (fitting) -- (infill);
    \path [l] (budget) -- node [font=\footnotesize] {yes} (return);
    \path [l] (budget) -- node [above, font=\footnotesize] {no} (fitting);
    \path [l] (infill) -- (update);
    \path [l] (update) -- (budget);
\end{tikzpicture}
  \end{center}
  \caption{General SMBO approach.}\label{fig:smbo_approach}
\end{figure}

Let $\fx: \X \to \R$ be an arbitrary black-box function with a $d$-dimensional input domain $\X = \X_1 \times \X_2 \times \cdots \times \X_d$ and a deterministic output~$y$.
Each $\X_i$ ($i=1,\cdots,d$) can be either numeric and bounded (\ie $\X_i = [l_i, u_i] \subset \R$) or a finite set of $s$ categorical values ($\X_i = \left\{ v_{i1}, \ldots, v_{is} \right\}$).
Without loss of generality, we want to find the input~$\xv^\ast$ with
\[ \xv^\ast = \argmin_{\xv\in\X} f(\xv). \]
In the context of \MBO, we usually assume that $f$ is expensive to evaluate, hence the total number of function evaluations is limited by a budget.
At the heart of SMBO are so-called surrogate models~\fh which cheaply estimate the expensive black-box function~$f$ and which are iteratively updated and refined.
The general approach is illustrated in Figure~\ref{fig:smbo_approach}.
The figure outlines the following steps, whereas each step is explained in more detail in the following subsections:
\begin{enumerate}[(1)]
  \item An initial design of $n_{\text{init}}$~points $\xv^{(j)}$ ($j = 1, \ldots, n_{\text{init}}$) is sampled from $\X$ and $f$ is evaluated at these points to yield outcomes $y^{(j)} = f(\xv^{(j)})$.
    The tuples $\left(y^{(j)}, \xv^{(j)}\right)$ constitute the data to build the initial surrogate model $\fh$ in the next step.
  \item \label{step:fit} Fit a surrogate model to all evaluated points $\xv^{(j)} \in \X$ and corresponding values $y^{(j)}$.
  \item An \emph{infill criterion} proposes $m$ points $\xv^{(j+i)}$ ($i = 1, \ldots, m$).
    The criterion is defined on \X and operates on the surrogate~\fh to
    determine points which are promising for the optimization.
    These points should either have a good expected objective value or high potential to improve the quality of the surrogate model.
  \item The proposed points are evaluated using $f$ and the new tuples $\left(y^{(j+i)}, \xv^{(j+i)}\right)$ are added to the design.
  \item If the budget is not exhausted (and no other termination criteria is met), go to step~\ref{step:fit}.
  \item If the budget is exhausted or another termination criteria is met, return the proposed solution for the optimization problem.
\end{enumerate}

\subsubsection{Initial Design}
\label{sssec:initial}
The initial design specifies the points of the input domain at which the
black-box function is evaluated to build the initial surrogate model~$\fh$.
If too few points are chosen or if the points do not cover $\X$ well, the fit of $\fh$ may be poor and thus points proposed based on $\fh$ may be suboptimal for the progress of the optimization.
Fitting a surrogate model may even be impossible.
On the other hand, a large initial design may reduce the available budget too
much.
\mlrMBO provides various options for the initial design:
The user can specify it manually or generate designs either completely at random, coarse grid designs or by using space-filling Latin Hypercube Designs~\citep{mckay1979comparison}.

\subsubsection{Surrogate Models}
\label{sssec:surrogate}
One of the main factors that determines the choice of surrogate model~$\fh$ is the structure of the input space \X.
If $\X \subset \R^d$, \emph{Kriging}~\cite{jones_1998} is the recommended choice and provides state-of-the-art performance.
In Section~\ref{ssec:ego}, the Kriging-based EGO approach is discussed in more detail.
If the search space \X also includes categorical parameters on the other hand, \emph{random forests} are a viable alternative~\cite{hutter_sequential_2011} as they can handle such parameters directly, without the need to encode the categorical parameters as numeric.
\mlrMBO allows the use of any of the many regression models available in the \Rlang package \mlr, which itself can also be easily extended to support custom regression learners~\cite{mlrtut_2016}.

While Kriging models and random forests already provide uncertainty estimation natively, generic bagging can be applied to arbitrary regression models to retrieve standard error estimators in \mlr.

\subsubsection{Infill Criteria}
\label{sssec:infill}
The infill criterion, or sometimes called acquisition function, guides the optimization and tries to trade-off exploitation and exploration.
This is usually achieved by combining \mhx and \shx (or \vhx) in a single formula in a well-balanced fashion, where the posterior mean \mhx and posterior standard deviation \shx (or posterior variance \vhx) are estimated by the surrogate model \fh.
\shx and \vhx are sometimes also called \enquote{local uncertainty estimators}.
Assuming that our model \fh is somewhat \enquote{spatial} in the sense that higher values of \shx indicate regions of the search space that few of our
design points lie close to and / or we have not learned the structure of $f$ very well at \xv, we are therefore looking for points with low \mhx and high \shx.

Arguably the most popular choice is the \emph{expected improvement}
\[ \EI(\xv) := \operatorname{E}(I(\xv)) \]
where the random variable $I(\xv)$ defines the potential improvement at $\xv$ over the currently best observed function value $y_{\min}$:
\[
  I(\xv) := \max\left\{y_{\min} - Y(\xv), 0\right\}.
\]
Here, $Y(\xv)$ is a random variable that should express the posterior distribution at $\xv$, estimated with \fh.
For a Gaussian process, $Y(\xv)$ is normally distributed with $Y(\xv) \sim N(\mhx, \vhx)$.
Under this assumption, $\EI(\xv)$ can be expressed analytically in closed form as
\begin{equation}
  \label{eq:ei}
  \EI(\xv) = \left( y_{\min} - \mhx \right) \Phi \left( \frac{y_{\min} - \mhx)}{\shx} \right) + \shx \phi \left( \frac{y_{\min} - \mhx}{\shx} \right),
\end{equation}
where $\Phi$ and $\phi$ are the distribution and density function of the standard normal distribution, respectively.

A simpler approach to balance \mhx and \shx for a point \xv is given by the \emph{lower confidence bound}
\begin{equation}
  \operatorname{LCB}(\xv, \lambda) = \mhx - \lambda \shx,
\end{equation}
where $\lambda > 0$ is a constant that controls the \enquote{mean vs. uncertainty} trade-off.

Furthermore, \mlrMBO currently support pure mean \mhx minimization (pure exploitation) and pure uncertainty \shx maximization (pure exploration) and further criteria for multiple point proposals (see Section \ref{ssec:multipoint}),  noisy optimization (see Section \ref{ssec:noisy}), and multi-objective optimization (See section \ref{ssec:multiobjective}).

\subsubsection{Infill Optimization}
\label{sssec:infill_opt}
The infill optimizer searches for the point~\xv which yields the best infill value.
Unlike the original optimization problem on $f$, the optimization on the infill criterion can be considered inexpensive.
While this is still a black-box optimization problem, points can be evaluated more lavishly, and~\citet{jones_1998} propose a branch and bound algorithm for this task.
\mlrMBO defaults to a more generic approach, which we call \emph{focus search}, outlined in Algorithm~\ref{algo:focus}.
It is able to handle numeric parameter spaces, categorical parameter spaces, as well as mixed and hierarchical spaces.
The algorithm starts with a large random design from which all points are evaluated by the surrogate regression model to determine the most promising point.
Next, focus search shrinks the search space around the best point and samples new random points for the now focused search space.
The shrinkage of search space is iterated $n_\text{iters}$ times.
The complete procedure can be restarted $n_\text{restart}$ times to avoid local optima.
Finally the best point over all restarts and iterations is returned.
Evolutionary algorithms like CMA-ES~\cite{R:cmaesr} or custom user-defined optimizers can be selected alternatively.

\begin{algorithm}[ht]
  \begin{algorithmic}[1]
    \Require infill criterion $c: \X \rightarrow \R$, control parameters $n_\text{restart}$, $n_\text{iters}$, $n_\text{points}$
    \For{$u \in \{1, ..., n_\text{restart}\}$}
     \State Set $\tilde{\X} = \X$
     \For{$v \in \{1, ..., n_\text{iters}\}$}
     \State generate random design $\mathcal D \subset \tilde{\X}$ of size $n_\text{points}$
         \State compute $\xv_{u, v}^* = (x_1^*, ..., x_d^*) = \arg\min_{\xv \in \mathcal D} c(\xv)$
         \State shrink $\tilde{\mathcal X}$ by focusing on $\mathbf x^*$:
         \For{each search space dimension $\tilde {\X}_i$ in $\tilde{\X}$}
           \If {$\tilde{\X}_i$ numeric: $\tilde{\X}_i = [l_i, u_i]$}
           \State $l_i = \max \{l_i, x_i^* - \frac{1}{4} (u_i - l_i)\}$
           \State $u_i = \min \{u_i, x_i^* + \frac{1}{4} (u_i - l_i)\} $
         \EndIf
       \If {$\tilde{\mathcal X}_i$ categorical: $\tilde{\mathcal X}_i = \left\{ v_{i1}, \ldots, v_{is} \right\}$, $s > 2$}
             \State $\bar{x}_i$ = sample one category uniformly from $\tilde{\X}_i \backslash x_i^*$
             \State $\tilde{\mathcal X}_i = \tilde{\mathcal X}_i \backslash \bar{x}_i$
       \EndIf
         \EndFor
       \EndFor
     \EndFor
     \State \textbf{Return} $\xv^* = \argmin\limits_{u \in \{1, ..., n_\text{restart}\}, v \in \{1, ..., n_\text{iters}\}} c(\xv_{u, v}^*)$
  \end{algorithmic}
  \caption{Infill Optimization: Focus Search.}
  \label{algo:focus}
\end{algorithm}

\subsubsection{Termination}
\label{sssec:termination}
Multiple termination criteria can be used in \mlrMBO.
Commonly a limit is set for the total number of evaluations of $f$ or for the number of SMBO iterations.
Alternatively, the optimization can be terminated after a given time or after a time budget for function evaluations is exhausted.
The optimization can also be stopped as soon as a predefined objective value is reached.
Furthermore, the user can create custom termination rules.

\subsubsection{Final Point}
\label{sssec:final_point}
Finally, the final solution $\xv^*$ has to be determined.
Usually the best point observed during the optimization is picked.
Fitting a last surrogate model to find the best point predicted is a viable option, especially if $f$ is noisy.

\subsection{Efficient Global Optimization (EGO)}
\label{ssec:ego}
Kriging models~\cite{RW06} are arguable the most popular choice for a surrogate model because they are very flexible and provide a local uncertainty estimator~\cite{jones_1998}.

In general, we consider a numeric-only input domain $\X \subset \R^d$.
\citet{jones_1998} were the first who introduced surrogate models for the sequential optimization of box-constrained functions with real-valued arguments.
Their Efficient Global Optimization (EGO) algorithm employs Kriging models together with the expected improvement infill criterion (see Equation~\ref{eq:ei}).
Maximizing the \EI results in an infill criterion that balances exploitation of the model structure and exploration of regions with high uncertainty and has proven to be highly effective~\cite{jones_1998}.
It can ensure global convergence \cite{vazquez2010,Jon01} (which is somewhat unrealistic to expect under the usually tight budget constraints that exist for many expensive black-box optimization problems).

\begin{figure}[hbt]
  \centering
\begin{knitrout}
\definecolor{shadecolor}{rgb}{0.969, 0.969, 0.969}\color{fgcolor}
\includegraphics[width=0.48\linewidth]{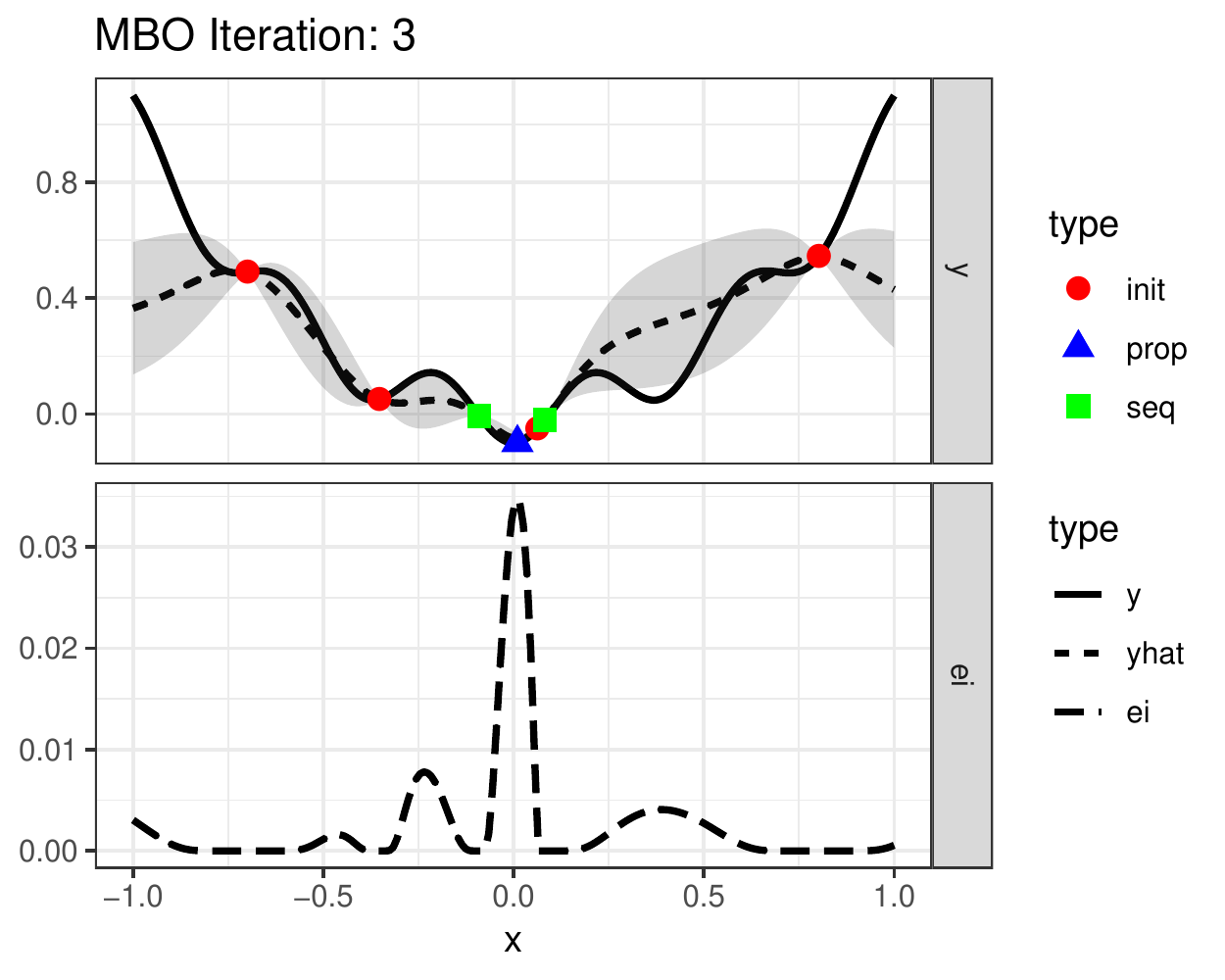}
\includegraphics[width=0.48\linewidth]{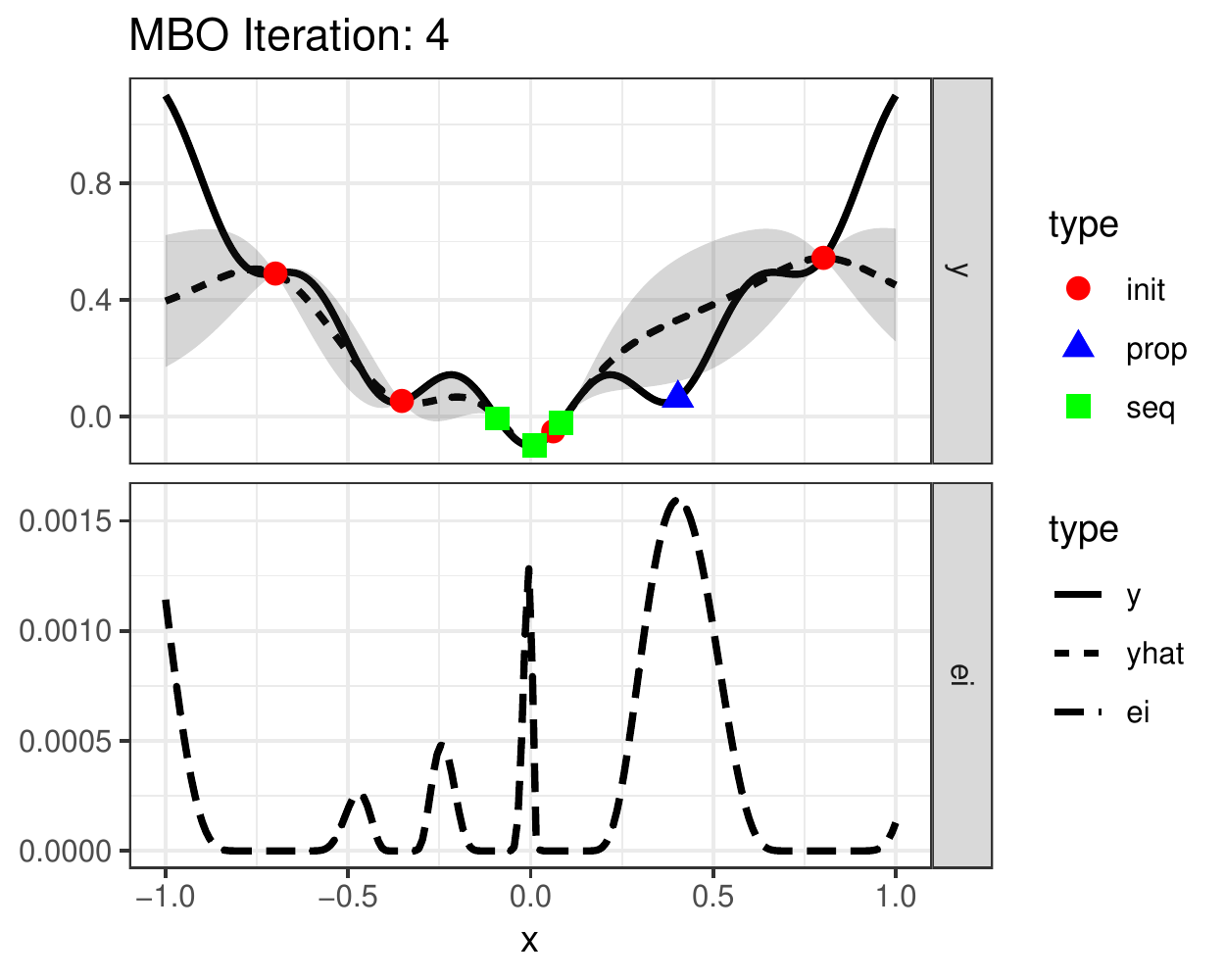}

\end{knitrout}
  \caption{State at the 3rd (left) and 4th iteration (right) of an exemplary EGO run on a $1d$ cosine mixture function.
  The upper part shows the real function $f$ as a solid line and its estimation~$\mh$ dotted.
  The uncertainty is indicated by the shaded area.
  Initial design points are displayed as red circles, sequential points as green squares.
  The lower part shows the respective value for the \EI.
  The optimum of the \EI defines the point that proposed to be evaluated next (blue triangle).}
  \label{fig:ego_example}
\end{figure}

Figure~\ref{fig:ego_example} illustrates the point proposal at the 3rd (left) and  the 4th iteration (right) of an EGO run on a $1d$ cosine mixture function.
It illustrates how high uncertainty ($\sh$) and a low value of $\mh$ contribute to the \EI and thus to the selection of the next point and the ability of \MBO to find the optimum even for multi-modal functions.

\subsection{Mixed Space Optimization}
\label{ssec:mso}

\begin{figure}[b]
  \center
  \begin{tikzpicture}[->,>=stealth',shorten >=1pt,auto,node distance=3cm,
    circ/.style={circle,draw,font=\scriptsize},
    rect/.style={rectangle,draw,font=\scriptsize}]
    \node[draw=none,fill=none] (20) at (0, 0) {$\mathcal X$};
    \node[circ]  (8) at (1, -1) {$C$};
    \node[circ]  (9) at(1, 1) {kernel};
    \node[rect] (17) at (2.5, 0.5){radial};
    \node[rect] (10) at (2.5, 1.5){linear};
    \node[circ] (11) at(4, 0.5) {$\gamma$};
    \node[rect] (15) at (5.5, -1) {$[0, \Inf]$};
    \node[rect] (16) at (5.5, 0.5) {$[0, \Inf]$};
    \path[every node/.style={font=\sffamily\small}]
    (20) edge node {}(8)
    edge node {}(9)
    (8) edge node {}(15)
    (9) edge node {}(10)
    edge node {}(17)
    (17) edge node {}(11)
    (11) edge node {}(16);
  \end{tikzpicture}
  \caption{Dependent search space for the tuning of a support vector machine. Circles denote parameters, rectangles denote parameter ranges, arrows denote the hierarchical structure.}
  \label{fig:svm_hyperparameters}
\end{figure}
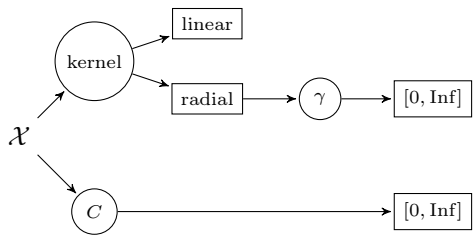

Real life scenarios often include mixed-valued as well as hierarchical parameter spaces with conditional parameters.
An example is the tuning of a support vector machine, for which the parameter space is illustrated in Figure~\ref{fig:svm_hyperparameters}.
Depending on the choice of the \emph{kernel}, the hyperparameter $\gamma$ has to be optimized for the \emph{radial} kernel (so it is conditional on the setting of \emph{kernel}), but it is not present (or we could say: active) for the \emph{linear} kernel.
In contrast to $\gamma$, the hyperparameter~$C$ is unconditionally always active.
Kriging is not really suited for such problem domains, since covariance kernels natively supporting those types of data are still subject to research~\cite{zhou2012simple}.

For the \emph{initial design} all options support categorical parameters as well as hierarchical dependencies (feasible values of a parameter depend on the values of other parameters).

For the \emph{surrogate} we need a regression model that is more flexible and can handle categorical features as well as missing values to support dependent parameters.
A slightly modified \emph{random forest} can be used for this purpose.
If a hyperparameter is not active in a design point in the training set (due to unfulfilled conditions), we will mark its value as missing.
Although the random forest could potentially directly handle missing values, many implementations do not.
Hence, we impute these values in the following way:
For categorical parameters we code missing values as a new level, and for numerical parameters we code the imputed value out of the range of the box-constraints of the parameter under consideration.
This is known as the \emph{separate-class method} and was shown to perform best for decision trees in a prediction-oriented study, when missingness is related to the outcome~\cite{ding_investigation_2010}.

In order to still use \emph{infill criteria} as LCB and EI, we also have to compute an uncertainty estimate \shx for the random forest.
For bagging-like predictors this can be computed or approximated in various ways from the bootstrap.
We refer the reader to \cite{sexton_rf_se_2009,wager_confidence_2014} for further details.
In \mlr the uncertainty estimator can be deviated from an expensive extra bootstrap around the random forest, the jackknife, the infinitesimal jackknife, or a simple estimator which extracts the standard error simply from the internal bootstrap of the random forest.
In our experience, the jackknife estimator works most reliably, so it is the current default for \mlrMBO with random forests as surrogate.
However, it should be noted that the random forest is not really a spatial model as a Gaussian process and therefore the properties of the uncertainty estimator are less intuitive in comparison to the ones from Kriging models.
Our following results still indicate that we obtained state-of-the-art results with this default, and we deem this aspect a matter for further research.

\subsection{Multi-Point Proposal}
\label{ssec:multipoint}
The expensive nature of the optimization problem makes parallelization, i.e. the evaluation of different configurations on multiple CPUs, an important extension to speed up the SMBO process.
Recently many methods have been proposed to simultaneously propose $m$ points in each iteration.
We showcase three methods implemented in \mlrMBO, which are also discussed in~\cite{bischl_2014}.
A straightforward approach is \emph{qLCB} \cite{hutter_2012}, an extension of the LCB criterion.
Instead of one fixed $\lambda$, multiple $\lambda_k\; (k = 1, \ldots, m)$ are drawn from an exponential distribution to obtain $m$ points $\xv^{(j+1)}, \ldots, \xv^{(j+m)}$:
\[
  \operatorname{qLCB}(\xv, \lambda_k) = \yh(\xv) - \lambda_k \shx, \quad \lambda_k \sim \operatorname{Exp}\left(\frac{1}{\lambda}\right).
\]
The criterion is than optimized separately for every $\lambda_k$, so that overall $m$ points are proposed.
Proposals that were obtained by optimizing the qLCB for a low value of $\lambda_k$ exploit the model and are in proximity of the best found $y$ so far.
For high values of $\lambda_k$ the proposals will be of exploratory nature.
This ensures that in one SMBO iteration all proposals balance exploitation and exploration.

Another approach to propose multiple points using the expected improvement is known as \emph{constant liar} \cite{ginsbourger2010kriging}.
Here we obtain $\xv^{(j + 1)}$ in the same way as for the ordinary \EI.
To obtain $\xv^{(j + 2)}$ we assume that the evaluation at point $\xv^{(j + 1)}$ is done and update the surrogate model with a made up target value $y$.
Exemplary choices for the made up value are $\min(\yv)$, $\max(\yv)$, the mean $\bar{\yv}$, or the predicted posterior mean $\hat\mu(\xv^{(j + 1)})$ of the surrogate model.
The latter approach is also often referred to as \emph{kriging believer}.


\citet{bischl_2014} propose the multi-objective infill \MBO (MOIMBO) approach.
The posterior mean $\mhx$ and variance $\shx$ are not scalarized in a single function (as done by \EI or (q)LCB), instead a multi-objective optimization strategy (see Section~\ref{ssec:multiobjective}) is used to optimize them jointly and propose a whole set of optimal points.
To ensure that the points are diverse, a distance measure, e.g. the nearest neighbor distance, can be used as a third objective.

\subsection{Noisy Optimization}
\label{ssec:noisy}

Noisy optimization assumes that the objective function $f$ is stochastic.
Usually, one now faces the problem to optimize $\operatorname{E}[f(x)]$ instead of $f(x)$ and common strategies are intelligent repetition strategies
\cite{Preuss06considerationsof} or adapted infill criteria.
\mlrMBO currently only offers the latter (but of course the user can always opt to perform averaging in the objective function, e.g.\ by naively averaging over a constant number of repetitions himself).

A popular infill criterion for noisy functions is the \emph{expected quantile improvement}~\cite{picheny2013quantile} which is an extension of \EI.
Instead of looking for an improvement over best value observed so far (the $y_{\min}$ in the \EI formula), we exchange this with a so called \enquote{plug-in} value $q_{\min}$:
\begin{equation}
  \operatorname{EQI}(\xv) = \left(q_{\min} - q(\xv) \right) \Phi \left( \frac{q_{\min} - q(\xv)}{s_{q(\xv)}} \right) + s_{q(\xv)} \phi \left( \frac{q_{\min} - q(\xv)}{s_{q(\xv)}} \right),
\end{equation}
where $q_{\min}$ is the lowest $\beta$-quantile $q(\xv_i)$ for all previously evaluated points $\xv \in \{\xv^{1}, \ldots \xv^{n}\}$, and $\beta$ is a user control parameter for the EQI.
The estimated $\beta$-quantile at point \xv is given by $q(\xv) = \mhx + \Phi^{-1}(\beta)\shx$.
This implies that the criterion will be non-zero at already evaluated points allowing re-evaluations or evaluations very close to already evaluated design points to increase knowledge of promising points.

\mlrMBO offers also the so called \enquote{augmented expected improvement} and its modular design makes extensions towards further criteria functions straightforward.
For a further in-depth discussion of this topic we refer the reader to \cite{PWG13} and their benchmark for noisy MBO approaches.



\subsection{Model-Based Multi-Objective (MBMO) Optimization}
\label{ssec:multiobjective}
Multi-objective optimization problems are characterized by a set of target functions $f(\xv) = (f_1(\xv), \dots, f_k(\xv))$ which have to be optimized simultaneously.
Since there is no total order in $\R^k, \text{for } k \geq 2$, the concept of \emph{Pareto dominance} is used.
A point $\xv$ pareto-dominates another point $\tilde \xv$, $\xv \preceq \tilde \xv$, if $f_i(\xv) \leq f_i(\tilde \xv)$ for $i = 1, \ldots, k$ and $\exists\, j\, f_j(\xv) < f_j(\tilde \xv)$, i.e., $\xv$ needs to be as good as $\tilde\xv$ in each component and strictly better in at least one.
A point $\xv$ is said to be \emph{non-dominated} if it is not dominated by any other point.
The set $P = \{\xv \, | \, \nexists\, \tilde \xv \; \tilde \xv \preceq \xv\}$ of all non-dominated points is called the \emph{Pareto set}.
It contains all incomparable trade-off solutions.
In multi-objective optimization the goal is to approximate the Pareto set or the \emph{Pareto front} $f(P)$, \ie, the image of $P$ under $f$.

In recent years some approaches were published that generalize single-objective SMBO algorithms like EGO for the multi-objective case.
We distinguish between 3 different MBMO algorithm classes:
First, \emph{scalarization based algorithms} that use EGO to optimize a scalarized version of the black-box functions with random weights for the scalarization in each iteration.
Second, \emph{Pareto based algorithms} that fit individual models for each objective and perform multi-objective optimization of infill criteria on these models.
Third, \emph{direct indicator based algorithms} that also fit individual models, but perform a single objective optimization of an infill criterion aggregating all models.
\mlrMBO supports 4 different MBMO algorithms, covering all 3 classes: ParEGO \cite{Kno06} as scalarization based, MSPOT \cite{ZBN+13} as Pareto based, and both SMS-EGO \cite{PWBV08} and $\varepsilon$-EGO \cite{Wag13} as direct indicator based algorithms.

A much more detailed discussion of these methods, their multi-point variants, and what is currently implemented in \mlrMBO is given in \cite{HWB15}.

\section{mlrMBO R Package}
\label{sec:software}
We implemented the software package \mlrMBO for the statistical programming language \Rlang.
It is designed as a modular framework.
The individual components of \MBO such as the infill criterion or the stopping conditions (cf.\ Section~\ref{sec:methodology}) can easily be combined in a plug-and-play fashion to respect the specific characteristics of the optimization problem at hand.
In the following we give a short introduction of this process which is split into multiple steps.

\paragraph{Definition of the black-box function}
For the first step \mlrMBO relies on the \pkg{smoof} package~\cite{R:smoof} which provides a unified interface to work with black-box functions.
Many test functions that are frequently used to benchmark optimizers are already included.
Additionally, the package provides the functions \code{make\{Single,Multi\}\allowbreak{}ObjectiveFunction()} as constructors for custom test functions.
Mandatory arguments are the function itself, a name and a parameter set.
In the simplest case, the latter is defined by names and box constraints, which can be specified concisely using the \pkg{ParamHelpers} package.
For more complex settings, it is also possible to connect parameters with arbitrary transformation functions (e.g., to vary a parameter on the $\log$-scale) or declare dependencies between parameters.
The following listing gives an example for the definition of the black-box $f(\mathbf{x}) = (x_2 - 0.1x_1^2 + x_1 - 6)^2 + \cos(x_1)$ with $x_1 \in [-5, 10], x_2 \in [0, 15]$:

\begin{knitrout}
\definecolor{shadecolor}{rgb}{0.969, 0.969, 0.969}\color{fgcolor}\begin{kframe}
\begin{alltt}
\hlstd{fn} \hlkwb{=} \hlkwd{makeSingleObjectiveFunction}\hlstd{(}
  \hlkwc{name} \hlstd{=} \hlstr{"my_blackbox"}\hlstd{,}
  \hlkwc{fn} \hlstd{=} \hlkwa{function}\hlstd{(}\hlkwc{x}\hlstd{) (x[}\hlnum{2}\hlstd{]} \hlopt{-} \hlnum{0.1} \hlopt{*} \hlstd{x[}\hlnum{1}\hlstd{]}\hlopt{^}\hlnum{2} \hlopt{+} \hlstd{x[}\hlnum{1}\hlstd{]} \hlopt{-} \hlnum{6}\hlstd{)}\hlopt{^}\hlnum{2} \hlopt{+} \hlkwd{cos}\hlstd{(x[}\hlnum{1}\hlstd{]),}
  \hlkwc{par.set} \hlstd{=} \hlkwd{makeParamSet}\hlstd{(}
    \hlkwd{makeNumericParam}\hlstd{(}\hlstr{"x1"}\hlstd{,} \hlkwc{lower} \hlstd{=} \hlopt{-}\hlnum{5}\hlstd{,} \hlkwc{upper} \hlstd{=} \hlnum{10}\hlstd{),}
    \hlkwd{makeNumericParam}\hlstd{(}\hlstr{"x2"}\hlstd{,} \hlkwc{lower} \hlstd{=} \hlnum{0}\hlstd{,} \hlkwc{upper} \hlstd{=} \hlnum{15}\hlstd{)}
  \hlstd{)}
\hlstd{)}
\end{alltt}
\end{kframe}
\end{knitrout}

\paragraph{Definition of the Initial Design}
To specify the points to be evaluated to initialize the surrogate an initial design has to be specified.
It is recommended to use a Latin Hypercube Design by calling \code{generateDesign()} and passing the number of desired points.
If no design is given by the user, \mlrMBO will generate a \textit{maximin} Latin Hypercube Design of size $4$ times the number of the black-box function's parameters.

\paragraph{Definition of the surrogate regression model}
\mlrMBO builds up on the \code{mlr} package~\cite{mlr_2016}, which offers a unified interface for a plethora of machine learning methods in \Rlang.
For surrogate regression, Kriging (\code{makeLearner("regr.km")}) and random forests (\code{makeLearner("regr.randomForest")}) are popular choices, but other regression methods can be selected as well.
Keep in mind that if expected improvement or LCB is chosen as the infill criterion, the surrogate either has to provide an uncertainty estimator, or has to be combined with a bagging approach using the \code{makeBaggingWrapper()} in \mlr.
If no regression method is supplied by the user, the fallback is a Kriging model with a Matern-\nicefrac{3}{2} kernel and the \enquote{GENetic Optimization Using Derivatives} (genoud) fitting algorithm in a fully numeric setting, and a random forest with jackknife variance estimation otherwise.

\paragraph{Definition of the control flow}
Basic settings like the number of proposed points in each SMBO iteration or the error handling are set via \code{makeMBOControl()} which returns a base control object.
This object can be further extended to adjust the different component of the SMBO methodology.
\code{setMBOControlInfill()} adjusts the infill criterion and the infill criterion optimizer.
If the infill optimization is unspecified, \mlrMBO uses LCB as infill criterion with $\lambda = 1$ in a fully numeric setting and $\lambda = 2$ if at least one one discrete parameter is present.
To optimize the criterion, focus search with $n_\text{restarts = 3}$, $n_\text{iters} = 5$ and $n_\text{points} = 1000$ is used by default.
For multi-point proposals or multi-objective optimization, \code{setMBOControlMultiPoint()} and \code{setMBOControlMultiObj()} are used, respectively.
If multiple points are proposed, they can be evaluated simultaneously using different parallelization (i.e. multicore, sockets, and MPI) and high-performance computation systems (e.g., Slurm, LSF, OpenLava, TORQUE, or Docker Swarms) with the \Rlang packages \pkg{parallelMap} and \pkg{batchtools}~\cite{batchjobs}.
Finally, \code{setMBOControlTermination()} controls the termination criteria.

\paragraph{Putting it all together}
The actual optimization is finally started by calling the \code{mbo()} function with the (optional) initial design, the black-box function, the (optional) surrogate regression method, and the control object as arguments.
The following listing demonstrates an application of \mlrMBO to optimize our example black-box.
\begin{knitrout}
\definecolor{shadecolor}{rgb}{0.969, 0.969, 0.969}\color{fgcolor}\begin{kframe}
\begin{alltt}
\hlkwd{library}\hlstd{(mlrMBO)}

\hlcom{# Create initial random Latin Hypercube Design of 10 points}
\hlkwd{library}\hlstd{(lhs)} \hlcom{# for randomLHS}
\hlstd{des} \hlkwb{=} \hlkwd{generateDesign}\hlstd{(}\hlkwc{n} \hlstd{=} \hlnum{5L} \hlopt{*} \hlnum{2L}\hlstd{,} \hlkwd{getParamSet}\hlstd{(fn),} \hlkwc{fun} \hlstd{= randomLHS)}

\hlcom{# Specify kriging model with standard error estimation}
\hlstd{surrogate} \hlkwb{=} \hlkwd{makeLearner}\hlstd{(}\hlstr{"regr.km"}\hlstd{,} \hlkwc{predict.type} \hlstd{=} \hlstr{"se"}\hlstd{,}
  \hlkwc{covtype} \hlstd{=} \hlstr{"matern3_2"}\hlstd{)}

\hlcom{# Set general controls}
\hlstd{ctrl} \hlkwb{=} \hlkwd{makeMBOControl}\hlstd{()}
\hlstd{ctrl} \hlkwb{=} \hlkwd{setMBOControlTermination}\hlstd{(ctrl,} \hlkwc{iters} \hlstd{=} \hlnum{30L}\hlstd{)}
\hlstd{ctrl} \hlkwb{=} \hlkwd{setMBOControlInfill}\hlstd{(ctrl,} \hlkwc{crit} \hlstd{=} \hlkwd{makeMBOInfillCritEI}\hlstd{())}

\hlcom{# Start optimization}
\hlkwd{mbo}\hlstd{(fn, des, surrogate, ctrl)}
\end{alltt}
\end{kframe}
\end{knitrout}
The resulting object contains the full optimization path, with all $x$ and $y$ values, runtime of function evaluations, final state, potential error messages as well as optionally all fitted surrogate models.
Diagnostic visualizations of the optimization are available by calling \code{plot()} and for one and two dimensional input domains with single- or multi-objective targets, each step of the optimization process can be visualized by calling \code{exampleRun()} or \code{exampleRunMultiObj()}.
For instance, Figure~\ref{fig:ego_example} has been created with \code{exampleRun()} and \code{plotExampleRun()}.

\section{Benchmarks}
\label{sec:benchmarks}

In this section, the performance of \mlrMBO is evaluated on three extensive benchmarks.
First, we compare \mlrMBO against other black-box optimizers connected to \Rlang (Section \ref{ssec:bench_single_opt}), then against state-of-the-art optimizers that are not available in \Rlang through the optimization benchmark framework HPOlib~\cite{eggensperger2013towards} (Section \ref{ssec:SMBO_mixed}).
Finally, we perform a simulation study on multi-objective optimization problems (Section \ref{ssec:bench_multi_opt}).
All benchmarks were conducted using the \texttt{batchtools}~\cite{batchjobs} package for \Rlang.

\subsection{Model-Based Single-Objective Optimization}
\label{ssec:bench_single_opt}

We run our implementation on various single-objective optimization tasks and compare it with the three EGO implementations available in \Rlang: \pkg{DiceOptim}~\cite{RGD12}, \pkg{rBayesianOptimization}~\cite{R:rBayesianOptimization} and \pkg{SPOT}~\cite{beielstein_spot_2012}.
Additionally, to ensure that an EGO approach is suitable, we also consider a basic random search as well as the popular covariance matrix adaptation evolution strategy (CMA-ES) based on the \Rlang package \pkg{cmaesr}~\cite{R:cmaesr}.

\paragraph{Benchmarks}
The methods are evaluated on a set of six $5$-dimensional, continuous, and single-objective test functions: \textit{Alpine01}, \textit{Deflected Corrugated Spring}, \textit{Schwefel}, \textit{Ackley}, \textit{Griewank} and \textit{Rosenbrock}.
All are defined in the \Rlang-package \pkg{smoof} and have been subject to optimization benchmarks previously.

\paragraph{Setup}
For the initial design, the same pre-generated maximin Latin Hypercube design containing 25~points is used for \mlrMBO, \pkg{DiceOptim}, \pkg{SPOT} and the random search.
It was not possible to pass a user-defined initial design in \pkg{rBayesianOptimization} without provoking an error.
Instead, a random design of the same size is generated internally.
We allow each algorithm \num{200} sequential iterations.
Since CMA-ES as an evolutionary algorithm does not initialize with a design, it gets an additional budget of 25~iterations (in total $225$).
All algorithms are run in their default settings carefully chosen by the respective package authors.

\begin{figure}[ht]
  \centering
\begin{knitrout}
\definecolor{shadecolor}{rgb}{0.969, 0.969, 0.969}\color{fgcolor}
\includegraphics[width=\maxwidth]{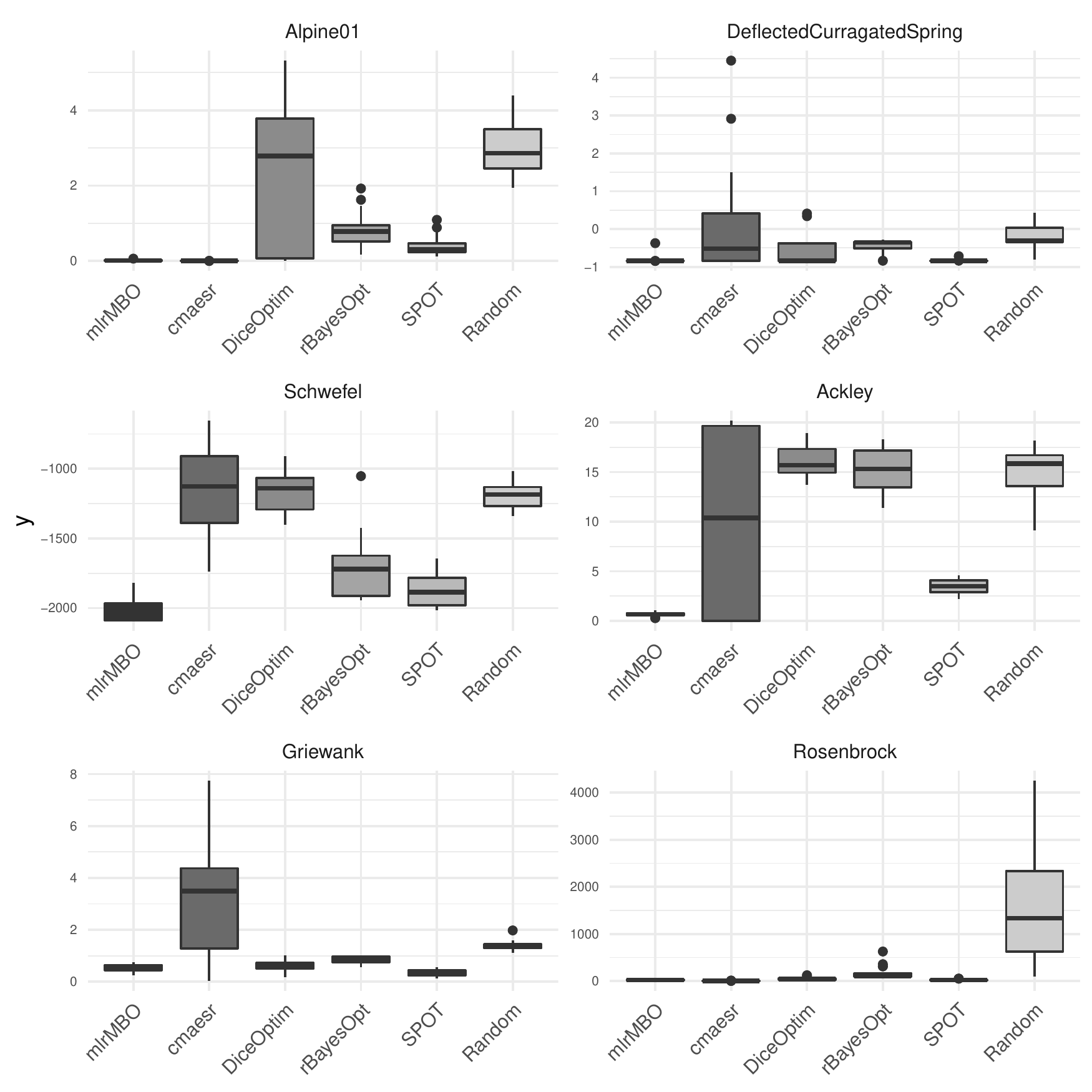}

\end{knitrout}
  \caption{Best objective value (on $y$ axis) found by respective algorithms on respective test function.}
  \label{fig:res_single_crit}
\end{figure}

\paragraph{Evaluation}
The objective values of the proposed solutions are summarized in Figure~\ref{fig:res_single_crit}.
All methods performed clearly better than the baseline random search approach on all six test functions.
In comparison with the other EGO-based algorithms, \mlrMBO yields a substantial better objective on four test functions and similar objective on the other two.
\pkg{SPOT} is slightly better than \mlrMBO on \textit{Griewank}, but worse on three others.
The evolutionary CMA-ES is comparable to \mlrMBO on \textit{Alpine01} and slightly better on \textit{Rosenbrock}, but considerably worse on the four other problems.
If we consider the averaged rank of the methods over all test functions as shown in Table~\ref{tab:testranks_syn}, \mlrMBO proves to be the best method overall, with \pkg{SPOT} in second place.

\begin{table}[ht]
\centering
\begin{tabular}{lrr}
  \hline
Algorithm & Average Rank & Average runtime in minutes \\
  \hline
\pkg{mlrMBO} & 1.95 & 8.03 \\
  \pkg{SPOT} & 2.48 & 27.88 \\
  \pkg{cmaesr} & 3.17 & 0.01 \\
  \pkg{DiceOptim} & 3.97 & 3.35 \\
  \pkg{rBayesOpt} & 4.24 & 695.96 \\
  \pkg{Random} & 5.19 & 0.00 \\
   \hline
\end{tabular}
\caption[Results were ranked in each replication and then averaged over the replications and problems. Runtime is measured in minutes.]{Average ranks and runtime on artificial test functions.}
\label{tab:testranks_syn}
\end{table}

Besides the quality of the solution, runtime, and computational overhead should also be considered.
The timings for a complete optimization run in minutes are listed in Table~\ref{tab:testranks_syn}.
Note that we are basically measuring the overhead of the optimization algorithms, as the synthetic test functions are evaluated in microseconds.
The random search unsurprisingly comes with the least overhead, followed by CMA-ES as implemented in the package \pkg{cmaesr}.
The EGO-based approaches consume considerably more time by fitting the surrogate model and optimizing the infill criterion.
Here, \mlrMBO is slower than \pkg{DiceOptim} but still more than twice as fast as \pkg{SPOT} and orders of magnitudes faster than \pkg{rBayesianOptimization}.
However, keep in mind that EGO is tailored for expensive problems.
If we paid each function evaluation with just one minute of computation time, the differences between \SI{200}{\minute} for random search and \SI{212}{\minute} for \mlrMBO seems to be a reasonable price to pay for a much better objective value.


\subsection{Model-Based Single-Objective Optimization in Mixed Spaces}
\label{ssec:SMBO_mixed}
The second benchmark compares \mlrMBO to three\footnote{Since \pkg{BayesOpt} is neither connected to HPOlib nor possesses an \Rlang interface, we refer the reader to the benchmarks in~\cite{JMLR:v15:martinezcantin14a} and do not consider \pkg{BayesOpt} in our analysis.} other state-of-the-art Bayesian optimizers which are not connected to \Rlang: Spearmint~\cite{snoek_practical_2012}, hyperopt (called TPE in the following)~\cite{bergstra2011algorithms} and SMAC~\cite{hutter_sequential_2011}.
We use the hyperparameter optimization library HPOlib~\cite{eggensperger2013towards}, which contains a large number of standardized benchmarks.
Besides purely numerical problems, the HPOlib also defines problems with mixed and dependent parameter spaces.
We evaluate the methods on four synthetic functions (\textit{branin}, \textit{camelback}, \textit{michalewicz} and \textit{har6}), three parameter optimization problems on grids (linear discriminant analysis (\textit{lda}), logistic regression (\textit{logreg}) and a support vector machine (\textit{svm})), as well as a deep neural network (\textit{hpnnet}) with $15$~parameters, and a deep belief network (\textit{hpdbnet}) with $35$~parameters.
The latter two problems were originally proposed by \citet{bergstra2011algorithms}. 
\mlrMBO uses its default settings, i.e., a Gaussian process as surrogate model for all solely numerical problems and a random forest for the problems with mixed and dependent parameter spaces (\textit{hpnnet} and \textit{hpdbnet}).
We deviate from the defaults only for the initial design of \textit{hpnnet} and \textit{hpdbnet}.
Here, the number of allowed function evaluations compared to the dimension of the parameter set is very small, therefore the initial design has only size $2d$ instead of the default $4d$.
Spearmint uses an internal dummy encoding of all categorical parameters for its Gaussian process.
The number of iterations on each benchmark as well as the specific settings of all other optimizers are defined in HPOlib.

\begin{figure}[ht!]
  \centering
\begin{knitrout}
\definecolor{shadecolor}{rgb}{0.969, 0.969, 0.969}\color{fgcolor}
\includegraphics[width=\maxwidth]{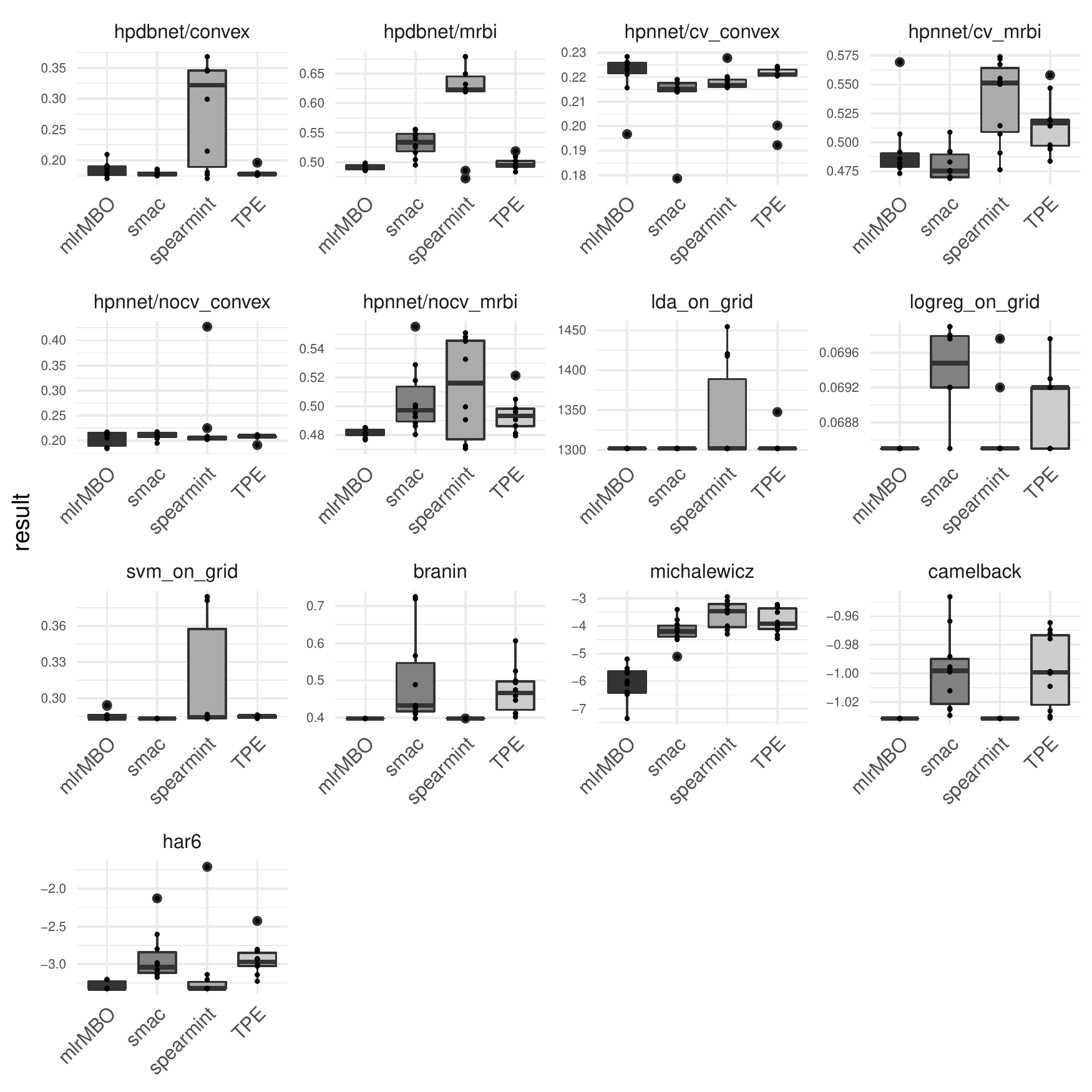}

\end{knitrout}
  \caption{Best objective value (on $y$ axis) found by respective optimizer on $13$ HPOlib test functions. For details on the test functions we refer to \cite{bergstra2011algorithms} for \textit{hpnnet} and \textit{hpdbnet} and to \cite{eggensperger2013towards} for everything else.}
  \label{fig:res_hpolib}
\end{figure}

\paragraph{Evaluation}
The results of ten runs on each benchmark are summarized in Figure~\ref{fig:res_hpolib}.
On each of the four synthetic test functions, \mlrMBO outperforms both SMAC and TPE and has similar performance compared to Spearmint, except for \textit{michalewicz} where \mlrMBO outperforms all competitors.
For the grid optimizations, \mlrMBO also performs exceptionally well on every single one, while each other optimizer results in worse performance on at least one of the three problems, which overall places \mlrMBO on the first place in numeric settings (cf.\ Table~\ref{tab:hporanks}).
Regarding the neural network and deep belief network, \mlrMBO achieves similar results as SMAC and slightly better results than Spearmint and TPE.
Especially Spearmint has a clearly worse performance on three of the six problems, while \mlrMBO is only worse than its competitors on \textit{hpdbnet/cv\_convex}.
As a result, Table~\ref{tab:hporanks} shows that \mlrMBO also places first w.r.t.\ aggregated mean ranks for mixed hyperparameter spaces.
We can clearly see that \mlrMBO is on par with other state-of-the-art Bayesian optimization software, even in highly complex settings.
\begin{table}[ht]
\centering
\begin{tabular}{llll}
  \hline
Optimizer & Avg. rank & Avg. rank (numeric only) & Avg. rank (mixed only) \\
  \hline
\pkg{mlrMBO} & 1.90 \textbf{(1)} & 1.64 \textbf{(1)} & 2.20 \textbf{(1)} \\
  \pkg{smac} & 2.65 \textbf{(3)} & 2.90 \textbf{(3)} & 2.35 \textbf{(2)} \\
  \pkg{spearmint} & 2.61 \textbf{(2)} & 2.32 \textbf{(2)} & 2.95 \textbf{(4)} \\
  \pkg{TPE} & 2.85 \textbf{(4)} & 3.14 \textbf{(4)} & 2.50 \textbf{(3)} \\
   \hline
\end{tabular}
\caption{Average ranks on HPOlib problems, Results were ranked in each replication and then averaged over the replications and problems. Numeric only ranks are based on benchmark $7$ to $13$ and mixed only ranks are based on $1$ to $6$.}
\label{tab:hporanks}
\end{table}

\subsection{Model-Based Multi-Objective Optimization}
\label{ssec:bench_multi_opt}
In \citet{HWB15} an exhaustive benchmark comparing all multi-objective algorithms implemented in \mlrMBO is given.
However, in 2018 an implementation bug that was likely to deteriorate the performance of ParEGO was discovered and fixed.
Therefore, we present a remake of the benchmark in this chapter, including a comparision to the \pkg{GPareto} \Rlang package.

\paragraph{Benchmarks}
The benchmark is performed on the bi-objective black-box optimization benchmarking (BBOB) test suite~\cite{tusar2016}.
It is constructed on top of ten functions of the single-objective BBOB test suite.
Two functions belong to each of the following function classes: separable (sep), moderate (mod), ill-condition (i-c), multi-modal (m-m) and weakly structured (w-s) functions.
These functions are pairwise combined to form 55 bi-objective problems, which can be grouped into 15 classes by combining the classes of the underlying single-objective functions.
The benchmark is restricted to the case $d = 5$.

\paragraph{Setup}
To simulate an expensive setting, all algorithms had a budget of $44d$~function evaluations, of which $4d$ are reserved for the initial design.
The popular evolutionary multi-objective algorithm NSGA2~\cite{DPAM02} and a random search serve as baseline for the four implementations in \mlrMBO: SMS-EGO, $\epsilon$-EGO, SMS-EGO, and MSPOT (cf.\ Section~\ref{ssec:multiobjective}).
In addition, the alternative implementation of SMS-EGO in \pkg{GPareto} is used.

\paragraph{Evaluation}
Various performance measures for comparing different approximations have been introduced.
The most popular measure may be the dominated hypervolume (also known as S-Metric).
In the bi-objective case the hypervolume simply measures the area between the discrete approximation of the Pareto front and a pessimistic reference point.
If an approximation reaches a higher hypervolume value, it is considered superior.

\begin{figure}[ht]
  \centering
\begin{knitrout}
\definecolor{shadecolor}{rgb}{0.969, 0.969, 0.969}\color{fgcolor}
\includegraphics[width=\maxwidth]{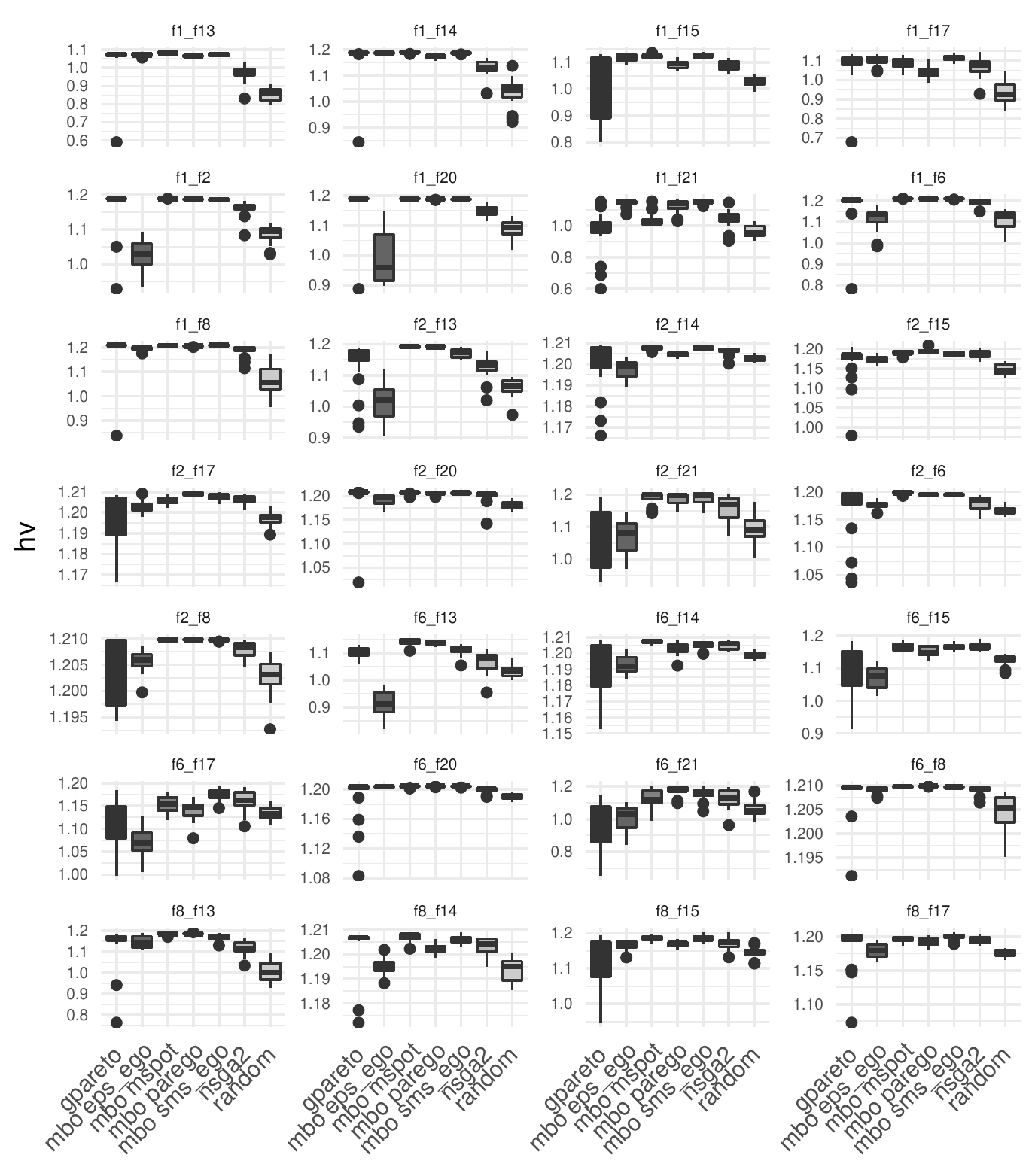}

\end{knitrout}
\caption{Hypervolume values of final Pareto fronts (on $y$ axis) found by respective algorithms on respective test function.}
\label{fig:res_multi_crit_hv_1}
\end{figure}

\begin{figure}[ht]
\centering
\begin{knitrout}
\definecolor{shadecolor}{rgb}{0.969, 0.969, 0.969}\color{fgcolor}
\includegraphics[width=\maxwidth]{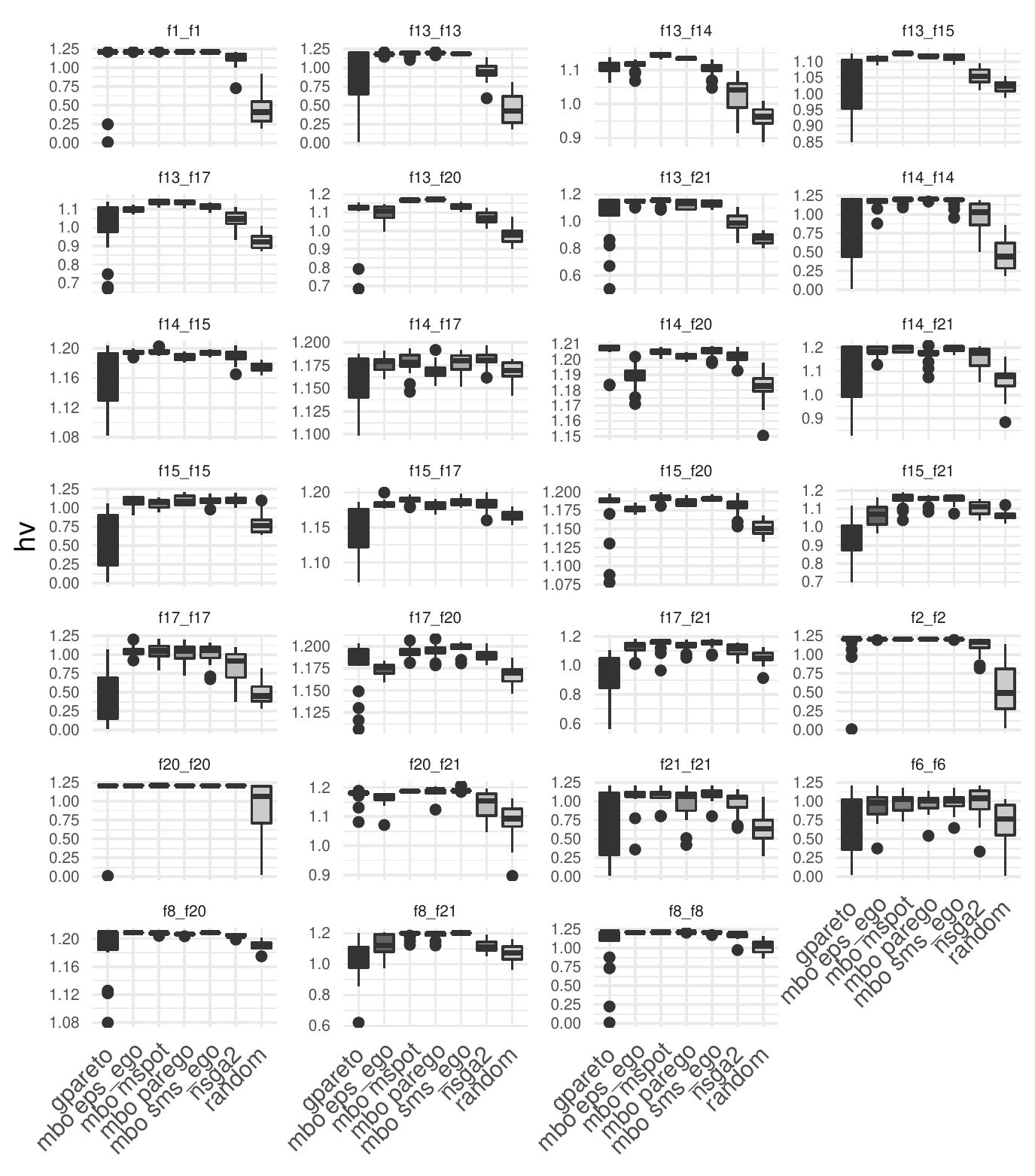}

\end{knitrout}
\caption{Hypervolume values of final Pareto fronts (on $y$ axis) found by respective algorithms on respective test function.}
\label{fig:res_multi_crit_hv_2}
\end{figure}

The final Pareto front approximations were normalized to the interval $[0, 1]^2$ with respect to a reference set.
Afterwards, ranks are computed for each test function respectively.
In Table~\ref{tab:multiranks}, the mean ranks for each function class are shown for 20~replications per test function.
Moreover, Figures~\ref{fig:res_multi_crit_hv_1} and \ref{fig:res_multi_crit_hv_2} show the raw hypervolume values for each test function.

We see that SMS-EGO, ParEGO and MSPOT outperform both baselines on nearly all test functions, with MSPOT beeing the superior algorithm.
Although gpareto has top performance for the class of two separable functions, it is inferior to all \mlrMBO implementation except $\epsilon$-EGO, especially while facing multi-modal functions.


\begin{table}[ht]
\begin{center}
\setlength{\tabcolsep}{4pt}
\begin{small}
\centering
\begin{tabular}{c|ccccccc}
\hline
group & GPareto & $\epsilon$-EGO & MSPOT & ParEGO & SMS-EGO & NSGA2 & random \\
  \hline
  sep -- sep & \textbf{2.18} & 4.72 & 2.22 & 2.88 & 3.50 & 5.30 & 6.40 \\
  sep -- mod & 3.45 & 5.55 & \textbf{1.85} & 2.56 & 2.61 & 4.85 & 6.46 \\
  sep -- i-c & 3.52 & 4.95 & \textbf{1.46} & 3.81 & 2.89 & 5.05 & 6.31 \\
  sep -- m-m & 4.42 & 3.96 & 3.09 & 3.39 & \textbf{2.51} & 3.90 & 6.72 \\
  sep -- w-s & 3.65 & 5.17 & 2.76 & 3.17 & \textbf{2.61} & 4.42 & 6.20 \\
  mod -- mod & 3.98 & 4.20 & 3.10 & \textbf{2.80} & 3.17 & 4.43 & 6.32 \\
  mod -- i-c & 3.81 & 5.95 & \textbf{1.68} & 2.90 & 3.11 & 4.40 & 6.15 \\
  mod -- m-m & 4.50 & 5.90 & 2.65 & 4.04 & \textbf{2.16} & 3.12 & 5.62 \\
  mod -- w-s & 4.96 & 3.91 & 2.73 & 2.91 & \textbf{2.38} & 4.83 & 6.14 \\
  i-c -- i-c & 4.22 & 3.63 & 2.22 & \textbf{2.07} & 3.50 & 5.65 & 6.72 \\
  i-c -- m-m & 5.16 & 3.45 & \textbf{2.16} & 3.54 & 3.17 & 4.39 & 6.12 \\
  i-c -- w-s & 3.67 & 4.24 & \textbf{2.41} & 3.19 & 2.92 & 4.89 & 6.67 \\
  m-m -- m-m & 6.12 & 3.13 & \textbf{2.85} & 3.22 & \textbf{2.85} & 3.55 & 6.28 \\
  m-m -- w-s & 5.26 & 4.90 & 2.24 & 3.08 & \textbf{2.23} & 4.19 & 6.11 \\
  w-s -- w-s & 4.17 & 3.98 & \textbf{2.88} & 3.12 & 2.53 & 4.68 & 6.63 \\ \hline
  over all   & 4.21 & 4.51 & \textbf{2.42} & 3.11 & 2.81 & 4.51 & 6.33 \\ \hline
\end{tabular}
\end{small}
\label{tab:multiranks}
\caption{Average ranks on bi-objective BBOB problems. Results were ranked in each replication and then averaged over the replications and problems for each function class.}
\end{center}
\end{table}

\section{Conclusion}
\label{sec:conclusion}
We introduced the \Rlang package \mlrMBO, a modular toolbox for \MBO in the \Rlang programming language.
We gave a brief introduction to software specific aspects and features.
Furthermore, we performed comprehensive benchmarks of \mlrMBO against other black-box optimizers in different scenarios.
In the single-objective benchmark \mlrMBO proved state-of-the-art performance regarding solution quality in comparison with the CMA evolutionary strategy, random search, and alternative SMBO implementations, while still being reasonably fast.
Furthermore, \mlrMBO is on par with the well known optimization frameworks SMAC, Spearmint, and TPE as shown by benchmarks using HPOlib.
The benchmark study on expensive multi-objective optimization revealed SMBO-based methods, in particular SMS-EGO, to show excellent performance.
Both the state-of-the-art NSGA-II evolutionary algorithm as well as the baseline random search algorithm were outperformed on all nine test functions (only ParEGO occasionally failed).
All in all the results demonstrate the suitability of the \mlrMBO toolbox in particular for expensive optimization scenarios in \Rlang for single- and multi-objective tasks, with continuous or mixed parameter spaces.


\section*{Acknowledgments}
Part of the work on this paper has been supported by Deutsche Forschungsgemeinschaft (DFG) within the Collaborative Research Center SFB 876 \enquote{Providing Information by Resource-Constrained Analysis}, project A3 (\href{http://sfb876.tu-dortmund.de}{http://sfb876.tu-dortmund.de}) and by the Competence Network for Technical, Scientific High Performance Computing in Bavaria (KONWIHR) in the project \enquote{Implementierung und Evaluation eines Verfahrens zur automatischen, massiv-parallelen Modellselektion im Maschinellen Lernen}.

\section*{References}
\bibliography{ms}

\begin{thebibliography}{46}
\providecommand{\natexlab}[1]{#1}
\providecommand{\url}[1]{\texttt{#1}}
\providecommand{\href}[2]{#2}
\providecommand{\path}[1]{#1}
\providecommand{\DOIprefix}{doi:}
\providecommand{\ArXivprefix}{arXiv:}
\providecommand{\URLprefix}{URL: }
\providecommand{\Pubmedprefix}{pmid:}
\providecommand{\doi}[1]{\href{http://dx.doi.org/#1}{\path{#1}}}
\providecommand{\Pubmed}[1]{\href{pmid:#1}{\path{#1}}}
\providecommand{\BIBand}{and}
\providecommand{\bibinfo}[2]{#2}
\ifx\xfnm\undefined \def\xfnm[#1]{\unskip,\space#1}\fi
\makeatletter\def\@biblabel#1{#1.}\makeatother
\bibitem[{Sieben et~al.(2010)Sieben, Wagner and Biermann}]{Sieben2010}
\bibinfo{author}{Sieben\xfnm[ B.]}, \bibinfo{author}{Wagner\xfnm[ T.]},
  \bibinfo{author}{Biermann\xfnm[ D.]}.
\newblock \bibinfo{title}{Empirical modeling of hard turning of aisi 6150 steel
  using design and analysis of computer experiments}.
\newblock \emph{\bibinfo{journal}{Production Engineering}}
  \bibinfo{year}{2010};\bibinfo{volume}{4}(\bibinfo{number}{2}):\bibinfo{pages}{115--125}.
\bibitem[{Sacks et~al.(1989)Sacks, Welch, Mitchell and Wynn}]{SWMW89}
\bibinfo{author}{Sacks\xfnm[ J.]}, \bibinfo{author}{Welch\xfnm[ W.J.]},
  \bibinfo{author}{Mitchell\xfnm[ T.J.]}, \bibinfo{author}{Wynn\xfnm[ H.P.]}.
\newblock \bibinfo{title}{Design and analysis of computer experiments}.
\newblock \emph{\bibinfo{journal}{Statistical science}}
  \bibinfo{year}{1989};\bibinfo{volume}{4}(\bibinfo{number}{4}):\bibinfo{pages}{409--423}.
\bibitem[{{Jones} et~al.(1998){Jones}, {Schonlau} and {Welch}}]{jones_1998}
\bibinfo{author}{{Jones}\xfnm[ D.R.]}, \bibinfo{author}{{Schonlau}\xfnm[ M.]},
  \bibinfo{author}{{Welch}\xfnm[ W.J.]}.
\newblock \bibinfo{title}{Efficient global optimization of expensive black-box
  functions}.
\newblock \emph{\bibinfo{journal}{Journal of {Global} {Optimization}}}
  \bibinfo{year}{1998};\bibinfo{volume}{13}(\bibinfo{number}{4}):\bibinfo{pages}{455--492}.
\bibitem[{Horn et~al.(2015)Horn, Wagner, Biermann, Weihs and Bischl}]{HWB15}
\bibinfo{author}{Horn\xfnm[ D.]}, \bibinfo{author}{Wagner\xfnm[ T.]},
  \bibinfo{author}{Biermann\xfnm[ D.]}, \bibinfo{author}{Weihs\xfnm[ C.]},
  \bibinfo{author}{Bischl\xfnm[ B.]}.
\newblock \bibinfo{title}{Model-based multi-objective optimization: Taxonomy,
  multi-point proposal, toolbox and benchmark}.
\newblock In: \bibinfo{editor}{Gaspar-Cunha\xfnm[ A.]},
  \bibinfo{editor}{Henggeler~Antunes\xfnm[ C.]}, \bibinfo{editor}{Coello\xfnm[
  C.C.]}, eds. \emph{\bibinfo{booktitle}{Evolutionary Multi-Criterion
  Optimization}}; vol. \bibinfo{volume}{9018} of \emph{\bibinfo{series}{Lecture
  Notes in Computer Science}}. \bibinfo{publisher}{Springer};
  \bibinfo{year}{2015}:\unskip \bibinfo{pages}{64--78}.
\bibitem[{Ginsbourger et~al.(2010)Ginsbourger, Le~Riche and
  Carraro}]{ginsbourger2010kriging}
\bibinfo{author}{Ginsbourger\xfnm[ D.]}, \bibinfo{author}{Le~Riche\xfnm[ R.]},
  \bibinfo{author}{Carraro\xfnm[ L.]}.
\newblock \bibinfo{title}{Kriging is well-suited to parallelize optimization}.
\newblock In: \emph{\bibinfo{booktitle}{Computational Intelligence in Expensive
  Optimization Problems}}. \bibinfo{publisher}{Springer};
  \bibinfo{year}{2010}:\unskip \bibinfo{pages}{131--162}.
\bibitem[{{Bischl} et~al.(2014){Bischl}, {Wessing}, {Bauer}, {Friedrichs} and
  {Weihs}}]{bischl_2014}
\bibinfo{author}{{Bischl}\xfnm[ B.]}, \bibinfo{author}{{Wessing}\xfnm[ S.]},
  \bibinfo{author}{{Bauer}\xfnm[ N.]}, \bibinfo{author}{{Friedrichs}\xfnm[
  K.]}, \bibinfo{author}{{Weihs}\xfnm[ C.]}.
\newblock \bibinfo{title}{{MOI}-{MBO}: Multiobjective infill for parallel
  model-based optimization}.
\newblock In: \emph{\bibinfo{booktitle}{Learning and {Intelligent}
  {Optimization} {Conference}}}. \bibinfo{year}{2014}:\unskip
  \bibinfo{pages}{173--186}.
\bibitem[{Hutter et~al.(2011)Hutter, Hoos and
  Leyton-Brown}]{hutter_sequential_2011}
\bibinfo{author}{Hutter\xfnm[ F.]}, \bibinfo{author}{Hoos\xfnm[ H.H.]},
  \bibinfo{author}{Leyton-Brown\xfnm[ K.]}.
\newblock \bibinfo{title}{Sequential model-based optimization for general
  algorithm configuration}.
\newblock In: \emph{\bibinfo{booktitle}{{LION} 5}}.
  \bibinfo{year}{2011}:\unskip \bibinfo{pages}{507--523}.
\bibitem[{Bergstra et~al.(2011)Bergstra, Bardenet, Bengio and
  K{\'e}gl}]{bergstra2011algorithms}
\bibinfo{author}{Bergstra\xfnm[ J.S.]}, \bibinfo{author}{Bardenet\xfnm[ R.]},
  \bibinfo{author}{Bengio\xfnm[ Y.]}, \bibinfo{author}{K{\'e}gl\xfnm[ B.]}.
\newblock \bibinfo{title}{Algorithms for hyper-parameter optimization}.
\newblock In: \emph{\bibinfo{booktitle}{Advances in Neural Information
  Processing Systems}}. \bibinfo{year}{2011}:\unskip
  \bibinfo{pages}{2546--2554}.
\bibitem[{{Thornton} et~al.(2013){Thornton}, {Hutter}, {Hoos} and
  {Leyton-Brown}}]{thornton_2013}
\bibinfo{author}{{Thornton}\xfnm[ C.]}, \bibinfo{author}{{Hutter}\xfnm[ F.]},
  \bibinfo{author}{{Hoos}\xfnm[ H.H.]}, \bibinfo{author}{{Leyton-Brown}\xfnm[
  K.]}.
\newblock \bibinfo{title}{Auto-{WEKA}: Combined selection and hyperparameter
  optimization of classification algorithms}.
\newblock In: \emph{\bibinfo{booktitle}{Proceedings of {ACM} {SIGKDD}}}.
  \bibinfo{year}{2013}:\unskip \bibinfo{pages}{847--855}.
\bibitem[{{Lang} et~al.(2015){Lang}, {Kotthaus}, {Marwedel}, {Weihs},
  {Rahnenf{\"u}hrer} and {Bischl}}]{lang_2015}
\bibinfo{author}{{Lang}\xfnm[ M.]}, \bibinfo{author}{{Kotthaus}\xfnm[ H.]},
  \bibinfo{author}{{Marwedel}\xfnm[ P.]}, \bibinfo{author}{{Weihs}\xfnm[ C.]},
  \bibinfo{author}{{Rahnenf{\"u}hrer}\xfnm[ J.]},
  \bibinfo{author}{{Bischl}\xfnm[ B.]}.
\newblock \bibinfo{title}{Automatic model selection for high-dimensional
  survival analysis}.
\newblock \emph{\bibinfo{journal}{Journal of {Statistical} {Computation} and
  {Simulation}}}
  \bibinfo{year}{2015};\bibinfo{volume}{85}(\bibinfo{number}{1}):\bibinfo{pages}{62--76}.
\bibitem[{Horn and Bischl(2016)}]{HB2016}
\bibinfo{author}{Horn\xfnm[ D.]}, \bibinfo{author}{Bischl\xfnm[ B.]}.
\newblock \bibinfo{title}{Multi-objective parameter configuration of machine
  learning algorithms using model-based optimization}.
\newblock In: \emph{\bibinfo{booktitle}{Computational Intelligence (SSCI), 2016
  IEEE Symposium Series on}}. \bibinfo{organization}{IEEE};
  \bibinfo{year}{2016}:\unskip \bibinfo{pages}{1--8}.
\bibitem[{Roustant et~al.(2012)Roustant, Ginsbourger and Deville}]{RGD12}
\bibinfo{author}{Roustant\xfnm[ O.]}, \bibinfo{author}{Ginsbourger\xfnm[ D.]},
  \bibinfo{author}{Deville\xfnm[ Y.]}.
\newblock \bibinfo{title}{{DiceKriging}, {DiceOptim}: Two {R} packages for the
  analysis of computer experiments by kriging-based metamodeling and
  optimization}.
\newblock \emph{\bibinfo{journal}{Journal of Statistical Software}}
  \bibinfo{year}{2012};\bibinfo{volume}{51}(\bibinfo{number}{1}):\bibinfo{pages}{1--55}.
\bibitem[{Yan(2016)}]{R:rBayesianOptimization}
\bibinfo{author}{Yan\xfnm[ Y.]}.
\newblock \bibinfo{title}{rBayesianOptimization: Bayesian Optimization of
  Hyperparameters}; \bibinfo{year}{2016}.
\newblock \URLprefix
  \url{https://CRAN.R-project.org/package=rBayesianOptimization};
  \bibinfo{note}{{R} package version 1.0.0}.
\bibitem[{Snoek et~al.(2012)Snoek, Larochelle and Adams}]{snoek_practical_2012}
\bibinfo{author}{Snoek\xfnm[ J.]}, \bibinfo{author}{Larochelle\xfnm[ H.]},
  \bibinfo{author}{Adams\xfnm[ R.P.]}.
\newblock \bibinfo{title}{Practical bayesian optimization of machine learning
  algorithms}.
\newblock In: \emph{\bibinfo{booktitle}{Advances in {Neural} {Information}
  {Processing} {Systems} 25}}. \bibinfo{publisher}{Curran Associates, Inc.};
  \bibinfo{year}{2012}:\unskip \bibinfo{pages}{2951--2959}.
\bibitem[{Hern{\'a}ndez-Lobato et~al.(2016)Hern{\'a}ndez-Lobato,
  Hern{\'a}ndez-Lobato, Shah and Adams}]{hernandez2016predictive}
\bibinfo{author}{Hern{\'a}ndez-Lobato\xfnm[ D.]},
  \bibinfo{author}{Hern{\'a}ndez-Lobato\xfnm[ J.M.]},
  \bibinfo{author}{Shah\xfnm[ A.]}, \bibinfo{author}{Adams\xfnm[ R.P.]}.
\newblock \bibinfo{title}{Predictive entropy search for multi-objective
  bayesian optimization}.
\newblock In: \emph{\bibinfo{booktitle}{Proceedings of the 33nd International
  Conference on Machine Learning (ICML)}}. \bibinfo{year}{2016}:\unskip
  \bibinfo{pages}{1492--1501}.
\bibitem[{Martinez-Cantin(2014)}]{JMLR:v15:martinezcantin14a}
\bibinfo{author}{Martinez-Cantin\xfnm[ R.]}.
\newblock \bibinfo{title}{Bayesopt: A bayesian optimization library for
  nonlinear optimization, experimental design and bandits}.
\newblock \emph{\bibinfo{journal}{Journal of Machine Learning Research}}
  \bibinfo{year}{2014};\bibinfo{volume}{15}:\bibinfo{pages}{3915--3919}.
\bibitem[{Bartz-Beielstein and Zaefferer(2012)}]{beielstein_spot_2012}
\bibinfo{author}{Bartz-Beielstein\xfnm[ T.]}, \bibinfo{author}{Zaefferer\xfnm[
  M.]}.
\newblock \bibinfo{title}{A gentle introduction to sequential parameter
  optimization}.
\newblock \bibinfo{type}{Tech. Rep.} \bibinfo{number}{2}; Bibliothek der
  Fachhochschule Koeln; \bibinfo{year}{2012}.
\newblock \URLprefix \url{http://opus.bsz-bw.de/fhk/volltexte/2012/19}.
\bibitem[{Bischl et~al.(2016)Bischl, Lang, Kotthoff, Schiffner, Richter,
  Studerus, Casalicchio and Jones}]{mlr_2016}
\bibinfo{author}{Bischl\xfnm[ B.]}, \bibinfo{author}{Lang\xfnm[ M.]},
  \bibinfo{author}{Kotthoff\xfnm[ L.]}, \bibinfo{author}{Schiffner\xfnm[ J.]},
  \bibinfo{author}{Richter\xfnm[ J.]}, \bibinfo{author}{Studerus\xfnm[ E.]},
  \bibinfo{author}{Casalicchio\xfnm[ G.]}, \bibinfo{author}{Jones\xfnm[ Z.M.]}.
\newblock \bibinfo{title}{{mlr}: Machine learning in {R}}.
\newblock \emph{\bibinfo{journal}{Journal of Machine Learning Research}}
  \bibinfo{year}{2016};\bibinfo{volume}{17}(\bibinfo{number}{170}):\bibinfo{pages}{1--5}.
\bibitem[{Koch et~al.(2012)Koch, Bischl, Flasch, Bartz-Beielstein, Weihs and
  Konen}]{koch2012}
\bibinfo{author}{Koch\xfnm[ P.]}, \bibinfo{author}{Bischl\xfnm[ B.]},
  \bibinfo{author}{Flasch\xfnm[ O.]}, \bibinfo{author}{Bartz-Beielstein\xfnm[
  T.]}, \bibinfo{author}{Weihs\xfnm[ C.]}, \bibinfo{author}{Konen\xfnm[ W.]}.
\newblock \bibinfo{title}{Tuning and evolution of support vector kernels}.
\newblock \emph{\bibinfo{journal}{Evolutionary Intelligence}}
  \bibinfo{year}{2012};\bibinfo{volume}{5}(\bibinfo{number}{3}):\bibinfo{pages}{153--170}.
\bibitem[{Bischl et~al.(2014)Bischl, Schiffner and Weihs}]{bischl2014}
\bibinfo{author}{Bischl\xfnm[ B.]}, \bibinfo{author}{Schiffner\xfnm[ J.]},
  \bibinfo{author}{Weihs\xfnm[ C.]}.
\newblock \bibinfo{title}{Benchmarking classification algorithms on
  high-performance computing clusters}.
\newblock In: \bibinfo{editor}{Spiliopoulou\xfnm[ M.]},
  \bibinfo{editor}{Schmidt-Thieme\xfnm[ L.]}, \bibinfo{editor}{Janning\xfnm[
  R.]}, eds. \emph{\bibinfo{booktitle}{Data Analysis, Machine Learning and
  Knowledge Discovery}}. Studies in Classification, Data Analysis, and
  Knowledge Organization; \bibinfo{publisher}{Springer};
  \bibinfo{year}{2014}:\unskip \bibinfo{pages}{23--31}.
\bibitem[{Hess et~al.(2013)Hess, Wagner and Bischl}]{hess2013}
\bibinfo{author}{Hess\xfnm[ S.]}, \bibinfo{author}{Wagner\xfnm[ T.]},
  \bibinfo{author}{Bischl\xfnm[ B.]}.
\newblock \bibinfo{title}{{PROGRESS}: Progressive reinforcement-learning-based
  surrogate selection}.
\newblock In: \bibinfo{editor}{Nicosia\xfnm[ G.]},
  \bibinfo{editor}{Pardalos\xfnm[ P.]}, eds. \emph{\bibinfo{booktitle}{Learning
  and Intelligent Optimization}}. Lecture Notes in Computer Science;
  \bibinfo{publisher}{Springer}; \bibinfo{year}{2013}:\unskip
  \bibinfo{pages}{110--124}.
\bibitem[{Horn et~al.(2016)Horn, Demircio{\u{g}}lu, Bischl, Glasmachers and
  Weihs}]{HDBGW2016}
\bibinfo{author}{Horn\xfnm[ D.]}, \bibinfo{author}{Demircio{\u{g}}lu\xfnm[
  A.]}, \bibinfo{author}{Bischl\xfnm[ B.]}, \bibinfo{author}{Glasmachers\xfnm[
  T.]}, \bibinfo{author}{Weihs\xfnm[ C.]}.
\newblock \bibinfo{title}{A comparative study on large scale kernelized support
  vector machines}.
\newblock \emph{\bibinfo{journal}{Advances in Data Analysis and
  Classification}} \bibinfo{year}{2016};:\bibinfo{pages}{1--17}.
\bibitem[{Steponavič et~al.(2016)Steponavič, Shirazi-Manesh, Hyndman,
  Smith-Miles and Villanova}]{Steponavice2016}
\bibinfo{author}{Steponavič\xfnm[ I.]}, \bibinfo{author}{Shirazi-Manesh\xfnm[
  M.]}, \bibinfo{author}{Hyndman\xfnm[ R.J.]},
  \bibinfo{author}{Smith-Miles\xfnm[ K.]}, \bibinfo{author}{Villanova\xfnm[
  L.]}.
\newblock \bibinfo{title}{On {Sampling} {Methods} for {Costly}
  {Multi}-{Objective} {Black}-{Box} {Optimization}}.
\newblock In: \bibinfo{editor}{Pardalos\xfnm[ P.M.]},
  \bibinfo{editor}{Zhigljavsky\xfnm[ A.]}, \bibinfo{editor}{Žilinskas\xfnm[
  J.]}, eds. \emph{\bibinfo{booktitle}{Advances in {Stochastic} and
  {Deterministic} {Global} {Optimization}}}. No. \bibinfo{number}{107} in
  \bibinfo{series}{Springer {Optimization} and {Its} {Applications}};
  \bibinfo{publisher}{Springer International Publishing}.
\newblock ISBN \bibinfo{isbn}{978-3-319-29973-0 978-3-319-29975-4};
  \bibinfo{year}{2016}:\unskip \bibinfo{pages}{273--296}.
\bibitem[{McKay et~al.(2000)McKay, Beckman and Conover}]{mckay1979comparison}
\bibinfo{author}{McKay\xfnm[ M.D.]}, \bibinfo{author}{Beckman\xfnm[ R.J.]},
  \bibinfo{author}{Conover\xfnm[ W.J.]}.
\newblock \bibinfo{title}{A comparison of three methods for selecting values of
  input variables in the analysis of output from a computer code}.
\newblock \emph{\bibinfo{journal}{Technometrics}}
  \bibinfo{year}{2000};\bibinfo{volume}{42}(\bibinfo{number}{1}):\bibinfo{pages}{55--61}.
\bibitem[{Schiffner et~al.(2016)Schiffner, Bischl, Lang, Richter, Jones,
  Probst, Pfisterer, Gallo, Kirchhoff, Kühn, Thomas and
  Kotthoff}]{mlrtut_2016}
\bibinfo{author}{Schiffner\xfnm[ J.]}, \bibinfo{author}{Bischl\xfnm[ B.]},
  \bibinfo{author}{Lang\xfnm[ M.]}, \bibinfo{author}{Richter\xfnm[ J.]},
  \bibinfo{author}{Jones\xfnm[ Z.M.]}, \bibinfo{author}{Probst\xfnm[ P.]},
  \bibinfo{author}{Pfisterer\xfnm[ F.]}, \bibinfo{author}{Gallo\xfnm[ M.]},
  \bibinfo{author}{Kirchhoff\xfnm[ D.]}, \bibinfo{author}{Kühn\xfnm[ T.]},
  \bibinfo{author}{Thomas\xfnm[ J.]}, \bibinfo{author}{Kotthoff\xfnm[ L.]}.
\newblock \bibinfo{title}{mlr tutorial}.
\newblock \bibinfo{year}{2016}.
\newblock \href{http://arxiv.org/abs/arXiv:1609.06146}{\tt
  arXiv:arXiv:1609.06146}.
\bibitem[{Bossek(2016)}]{R:cmaesr}
\bibinfo{author}{Bossek\xfnm[ J.]}.
\newblock \bibinfo{title}{cmaesr: Covariance Matrix Adaptation Evolution
  Strategy}; \bibinfo{year}{2016}.
\newblock \URLprefix \url{https://CRAN.R-project.org/package=cmaesr};
  \bibinfo{note}{{R} package version 1.0.1}.
\bibitem[{Rasmussen and Williams(2006)}]{RW06}
\bibinfo{author}{Rasmussen\xfnm[ C.E.]}, \bibinfo{author}{Williams\xfnm[
  C.K.I.]}.
\newblock \bibinfo{title}{Gaussian Processes for Machine Learning}.
\newblock Adaptive Computation and Machine Learning; \bibinfo{publisher}{MIT
  Press}; \bibinfo{year}{2006}.
\bibitem[{Vazquez and Bect(2010)}]{vazquez2010}
\bibinfo{author}{Vazquez\xfnm[ E.]}, \bibinfo{author}{Bect\xfnm[ J.]}.
\newblock \bibinfo{title}{Convergence properties of the expected improvement
  algorithm with fixed mean and covariance functions}.
\newblock \emph{\bibinfo{journal}{Journal of Statistical Planning and
  Inference}}
  \bibinfo{year}{2010};\bibinfo{volume}{140}(\bibinfo{number}{11}):\bibinfo{pages}{3088--3095}.
\bibitem[{Jones(2001)}]{Jon01}
\bibinfo{author}{Jones\xfnm[ D.R.]}.
\newblock \bibinfo{title}{{A taxonomy of global optimization methods based on
  response surfaces}}.
\newblock \emph{\bibinfo{journal}{Journal of Global Optimization}}
  \bibinfo{year}{2001};\bibinfo{volume}{21}(\bibinfo{number}{4}):\bibinfo{pages}{345--383}.
\bibitem[{Zhou et~al.(2012)Zhou, Qian and Zhou}]{zhou2012simple}
\bibinfo{author}{Zhou\xfnm[ Q.]}, \bibinfo{author}{Qian\xfnm[ P.Z.]},
  \bibinfo{author}{Zhou\xfnm[ S.]}.
\newblock \bibinfo{title}{A simple approach to emulation for computer models
  with qualitative and quantitative factors}.
\newblock \emph{\bibinfo{journal}{Technometrics}} \bibinfo{year}{2012};.
\bibitem[{Ding and Simonoff(2010)}]{ding_investigation_2010}
\bibinfo{author}{Ding\xfnm[ Y.]}, \bibinfo{author}{Simonoff\xfnm[ J.S.]}.
\newblock \bibinfo{title}{An investigation of missing data methods for
  classification trees applied to binary response data}.
\newblock \emph{\bibinfo{journal}{Journal of Machine Learning Research}}
  \bibinfo{year}{2010};\bibinfo{volume}{11}:\bibinfo{pages}{131--170}.
\newblock \URLprefix \url{http://www.jmlr.org/papers/v11/ding10a.html}.
\bibitem[{Sexton and Laake(2009)}]{sexton_rf_se_2009}
\bibinfo{author}{Sexton\xfnm[ J.]}, \bibinfo{author}{Laake\xfnm[ P.]}.
\newblock \bibinfo{title}{{Standard errors for bagged and random forest
  estimators}}.
\newblock \emph{\bibinfo{journal}{Computational Statistics \& Data Analysis}}
  \bibinfo{year}{2009};\bibinfo{volume}{53}(\bibinfo{number}{3}):\bibinfo{pages}{801--811}.
\bibitem[{Wager et~al.(2014)Wager, Hastie and Efron}]{wager_confidence_2014}
\bibinfo{author}{Wager\xfnm[ S.]}, \bibinfo{author}{Hastie\xfnm[ T.]},
  \bibinfo{author}{Efron\xfnm[ B.]}.
\newblock \bibinfo{title}{Confidence intervals for random forests: The
  jackknife and the infinitesimal jackknife}.
\newblock \emph{\bibinfo{journal}{Journal of Machine Learning Research}}
  \bibinfo{year}{2014};\bibinfo{volume}{15}:\bibinfo{pages}{1625--1651}.
\newblock \URLprefix \url{http://jmlr.org/papers/v15/wager14a.html}.
\bibitem[{Hutter et~al.(2012)Hutter, Hoos and Leyton-Brown}]{hutter_2012}
\bibinfo{author}{Hutter\xfnm[ F.]}, \bibinfo{author}{Hoos\xfnm[ H.H.]},
  \bibinfo{author}{Leyton-Brown\xfnm[ K.]}.
\newblock \bibinfo{title}{Parallel algorithm configuration}.
\newblock In: \emph{\bibinfo{booktitle}{Learning and Intelligent
  Optimization}}. \bibinfo{publisher}{Springer}; \bibinfo{year}{2012}:\unskip
  \bibinfo{pages}{55--70}.
\bibitem[{Preuss(2006)}]{Preuss06considerationsof}
\bibinfo{author}{Preuss\xfnm[ M.]}.
\newblock \bibinfo{title}{Considerations of budget allocation for sequential
  parameter optimization (spo}.
\newblock In: \emph{\bibinfo{booktitle}{Workshop on Empirical Methods for the
  Analysis of Algorithms, Proceedings}}. \bibinfo{year}{2006}:\unskip
  \bibinfo{pages}{35--40}.
\bibitem[{Picheny et~al.(2013{\natexlab{a}})Picheny, Ginsbourger, Richet and
  Caplin}]{picheny2013quantile}
\bibinfo{author}{Picheny\xfnm[ V.]}, \bibinfo{author}{Ginsbourger\xfnm[ D.]},
  \bibinfo{author}{Richet\xfnm[ Y.]}, \bibinfo{author}{Caplin\xfnm[ G.]}.
\newblock \bibinfo{title}{Quantile-based optimization of noisy computer
  experiments with tunable precision}.
\newblock \emph{\bibinfo{journal}{Technometrics}}
  \bibinfo{year}{2013}{\natexlab{a}};\bibinfo{volume}{55}(\bibinfo{number}{1}):\bibinfo{pages}{2--13}.
\bibitem[{Picheny et~al.(2013{\natexlab{b}})Picheny, Wagner and
  Ginsbourger}]{PWG13}
\bibinfo{author}{Picheny\xfnm[ V.]}, \bibinfo{author}{Wagner\xfnm[ T.]},
  \bibinfo{author}{Ginsbourger\xfnm[ D.]}.
\newblock \bibinfo{title}{{A benchmark of kriging-based infill criteria for
  noisy optimization}}.
\newblock \emph{\bibinfo{journal}{Struct Multidiscip Optim}}
  \bibinfo{year}{2013}{\natexlab{b}};\bibinfo{volume}{48}(\bibinfo{number}{3}):\bibinfo{pages}{607--626}.
\bibitem[{Knowles(2006)}]{Kno06}
\bibinfo{author}{Knowles\xfnm[ J.]}.
\newblock \bibinfo{title}{{ParEGO}: A hybrid algorithm with on-line landscape
  approximation for expensive multiobjective optimization problems}.
\newblock \emph{\bibinfo{journal}{IEEE Transactions on Evolutionary
  Computation}}
  \bibinfo{year}{2006};\bibinfo{volume}{10}(\bibinfo{number}{1}):\bibinfo{pages}{50--66}.
\bibitem[{Zaefferer et~al.(2013)Zaefferer, Bartz-Beielstein, Naujoks, Wagner
  and Emmerich}]{ZBN+13}
\bibinfo{author}{Zaefferer\xfnm[ M.]}, \bibinfo{author}{Bartz-Beielstein\xfnm[
  T.]}, \bibinfo{author}{Naujoks\xfnm[ B.]}, \bibinfo{author}{Wagner\xfnm[
  T.]}, \bibinfo{author}{Emmerich\xfnm[ M.]}.
\newblock \bibinfo{title}{A case study on multi-criteria optimization of an
  event detection software under limited budgets}.
\newblock In: \bibinfo{editor}{Purshouse\xfnm[ R.]}, et~al., eds.
  \emph{\bibinfo{booktitle}{Proc. 7th Int'l. Conf. Evolutionary Multi-Criterion
  Optimization (EMO)}}. \bibinfo{publisher}{Springer};
  \bibinfo{year}{2013}:\unskip \bibinfo{pages}{756--770}.
\bibitem[{Ponweiser et~al.(2008)Ponweiser, Wagner, Biermann and
  Vincze}]{PWBV08}
\bibinfo{author}{Ponweiser\xfnm[ W.]}, \bibinfo{author}{Wagner\xfnm[ T.]},
  \bibinfo{author}{Biermann\xfnm[ D.]}, \bibinfo{author}{Vincze\xfnm[ M.]}.
\newblock \bibinfo{title}{Multiobjective optimization on a limited amount of
  evaluations using model-assisted $\mathcal{S}$-metric selection}.
\newblock In: \emph{\bibinfo{booktitle}{Proc. 10th Int'l Conf. Parallel Problem
  Solving from Nature (PPSN)}}. \bibinfo{year}{2008}:\unskip
  \bibinfo{pages}{784--794}.
\bibitem[{Wagner(2013)}]{Wag13}
\bibinfo{author}{Wagner\xfnm[ T.]}.
\newblock \bibinfo{title}{Planning and Multi-Objective Optimization of
  Manufacturing Processes by Means of Empirical Surrogate Models}.
\newblock \bibinfo{publisher}{Vulkan Verlag, Essen}; \bibinfo{year}{2013}.
\bibitem[{Bossek(2017)}]{R:smoof}
\bibinfo{author}{Bossek\xfnm[ J.]}.
\newblock \bibinfo{title}{{smoof: Single- and Multi-Objective Optimization Test
  Functions}}.
\newblock \emph{\bibinfo{journal}{{The R Journal}}}
  \bibinfo{year}{2017};\URLprefix
  \url{https://journal.r-project.org/archive/2017/RJ-2017-004/index.html}.
\bibitem[{Bischl et~al.(2015)Bischl, Lang, Mersmann, Rahnenf{\"u}hrer and
  Weihs}]{batchjobs}
\bibinfo{author}{Bischl\xfnm[ B.]}, \bibinfo{author}{Lang\xfnm[ M.]},
  \bibinfo{author}{Mersmann\xfnm[ O.]}, \bibinfo{author}{Rahnenf{\"u}hrer\xfnm[
  J.]}, \bibinfo{author}{Weihs\xfnm[ C.]}.
\newblock \bibinfo{title}{{BatchJobs} and {BatchExperiments}: Abstraction
  mechanisms for using {R} in batch environments}.
\newblock \emph{\bibinfo{journal}{Journal of Statistical Software}}
  \bibinfo{year}{2015};\bibinfo{volume}{64}(\bibinfo{number}{11}):\bibinfo{pages}{1--25}.
\bibitem[{Eggensperger et~al.(2013)Eggensperger, Feurer, Hutter, Bergstra,
  Snoek, Hoos and Leyton-Brown}]{eggensperger2013towards}
\bibinfo{author}{Eggensperger\xfnm[ K.]}, \bibinfo{author}{Feurer\xfnm[ M.]},
  \bibinfo{author}{Hutter\xfnm[ F.]}, \bibinfo{author}{Bergstra\xfnm[ J.]},
  \bibinfo{author}{Snoek\xfnm[ J.]}, \bibinfo{author}{Hoos\xfnm[ H.]},
  \bibinfo{author}{Leyton-Brown\xfnm[ K.]}.
\newblock \bibinfo{title}{Towards an empirical foundation for assessing
  bayesian optimization of hyperparameters}.
\newblock In: \emph{\bibinfo{booktitle}{NIPS workshop on Bayesian Optimization
  in Theory and Practice}}. \bibinfo{year}{2013}:\unskip \bibinfo{pages}{1--5}.
\bibitem[{Tusar et~al.(2016)Tusar, Brockhoff, Hansen and Auger}]{tusar2016}
\bibinfo{author}{Tusar\xfnm[ T.]}, \bibinfo{author}{Brockhoff\xfnm[ D.]},
  \bibinfo{author}{Hansen\xfnm[ N.]}, \bibinfo{author}{Auger\xfnm[ A.]}.
\newblock \bibinfo{title}{{COCO:} the bi-objective black box optimization
  benchmarking (bbob-biobj) test suite}.
\newblock \emph{\bibinfo{journal}{CoRR}}
  \bibinfo{year}{2016};\bibinfo{volume}{abs/1604.00359}.
\newblock \URLprefix \url{http://arxiv.org/abs/1604.00359}.
\bibitem[{Deb et~al.(2002)Deb, Pratap, Agarwal and Meyarivan}]{DPAM02}
\bibinfo{author}{Deb\xfnm[ K.]}, \bibinfo{author}{Pratap\xfnm[ A.]},
  \bibinfo{author}{Agarwal\xfnm[ S.]}, \bibinfo{author}{Meyarivan\xfnm[ T.]}.
\newblock \bibinfo{title}{A fast and elitist multiobjective genetic algorithm:
  {NSGA-II}}.
\newblock \emph{\bibinfo{journal}{IEEE Transactions on Evolutionary
  Computation}}
  \bibinfo{year}{2002};\bibinfo{volume}{6}(\bibinfo{number}{2}):\bibinfo{pages}{182--197}.

\end{thebibliography}

\clearpage

\section*{Appendix}

\end{document}